\documentclass[sigconf, nonacm]{acmart}
\usepackage{xcolor}
\usepackage{xurl}
\usepackage[utf8]{inputenc}
\usepackage{graphicx} 
\usepackage{multirow}
\usepackage{rotating} 
\usepackage{multirow}
\usepackage{booktabs}
\usepackage{soul}
\usepackage{pifont}
\usepackage[htt]{hyphenat}
\usepackage{subcaption}
\usepackage{units}

\newcommand{\Smore}{S_{\text{more}}}
\newcommand{\Sless}{S_{\text{less}}}

\newcommand{\fmeval}{\texttt{fmeval}}

\AtBeginDocument{%
  \providecommand\BibTeX{{%
    \normalfont B\kern-0.5em{\scshape i\kern-0.25em b}\kern-0.8em\TeX}}}

\setcopyright{none}
\acmConference{}{}{}
\begin{document}

\title{Evaluating Large Language Models with \texttt{fmeval}}

\author{Pola Schw\"obel}
\authornote{The authors contributed equally to this work.}
\email{schwobel@amazon.de}
\orcid{0000-0003-4846-1917}
\affiliation{%
  \institution{Amazon Web Services}
  \city{Berlin}
  \country{Germany}
  }

\author{Luca Franceschi}
\authornotemark[1]
\email{franuluc@amazon.de}
\orcid{0000-0002-1810-1016}
\affiliation{%
  \institution{Amazon Web Services}
  \city{Berlin}
  \country{Germany}
}

\author{Muhammad Bilal Zafar}
\authornote{Work done while at Amazon.}
\orcid{0000-0003-4846-1917}
\affiliation{%
  \institution{Ruhr University Bochum}
  \city{Bochum}
  \country{Germany}
}

\author{Keerthan Vasist}
\affiliation{%
  \institution{Amazon Web Services}
  \city{Santa Clara}
  \country{USA}
}

\author{Aman Malhotra}
\affiliation{%
  \institution{Amazon Web Services}
  \city{Santa Clara}
  \country{USA}
}

\author{Tomer Shenhar}
\affiliation{%
  \institution{Amazon Web Services}
  \city{New York}
  \country{USA}
}

\author{Pinal Tailor}
\affiliation{%
  \institution{Amazon Web Services}
  \city{Arlington}
  \country{USA}
}

\author{Pinar Yilmaz}
\affiliation{%
  \institution{Amazon Web Services}
  \city{Santa Clara}
  \country{USA}
}

\author{Michael Diamond}
\affiliation{%
  \institution{Amazon Web Services}
  \city{Santa Clara}
  \country{USA}
}

\author{Michele Donini}
\orcid{0000-0002-9769-3899}
\affiliation{%
  \institution{Amazon Web Services}
  \city{Berlin}
  \country{Germany}
}

\renewcommand{\shortauthors}{Schw\"obel and Franceschi, et al.}

\begin{abstract}
\texttt{fmeval} is an open source library to evaluate large language models (LLMs) in a range of tasks. It helps practitioners evaluate their model for task performance and along multiple responsible AI dimensions. This paper presents the library and exposes its underlying design principles: simplicity, coverage, extensibility and performance. We then present how these were implemented in the scientific and engineering choices taken when developing \texttt{fmeval}. A case study demonstrates a typical use case for the library: picking a suitable model for a question answering task. We close by discussing limitations and further work in the development of the library. \texttt{fmeval} can be found at \url{https://github.com/aws/fmeval}.

\end{abstract}

\maketitle

\pagestyle{empty}

\section{Introduction}
The advent of foundation models (FMs) as the workhorse for generative AI has revolutionized machine learning (ML). The potential for automation on an unprecedented scale promises efficiency leaps in a wide range of industries, such as finance, healthcare, public service, travel and hospitality.
Meanwhile, the risks associated with generative AI have been well-publicized, especially for models in the language domain \cite{bender2021dangers, blodgett2020language, sheng2021societal, wang2023decodingtrust}.
Large language models (LLMs) are trained on volumes of data, including undesirable content laden with historical biases, undemocratic viewpoints or hate speech. As a consequence, LLMs are at risk of regurgitating toxicity \cite{gehman2020realtoxicityprompts, dhamala2021bold}, stereotypes \cite{nadeem2020stereoset, nangia2020crows, nozza2021honest, abid2021persistent}, and non-truthful outputs \cite{lin2021truthfulqa, ji2023survey}. Such model behaviors can cause harm to users, damage an organization’s reputation and jeopardize customer trust. Additionally, ethical and safety dimensions of ML models have recently been under close regulatory scrutiny via guidelines and regulations, such as ISO 42001 or the EU AI Act.

Detecting and managing risks, as prescribed by such guidelines, is challenging. ML engineers and data scientists have to leave their development environment to use academic tools and benchmarking sites, which require highly-specialized knowledge. The sheer number of metrics makes it hard to identify the ones that are truly relevant for their use-cases, while evaluating all of them leads to high compute costs and can take several days to run for a single model. This tedious process then needs to be repeated frequently as new models are released and existing ones are fine-tuned.

Simplifying this process, \texttt{fmeval} provides users a single place to evaluate and compare metrics during the model selection and model customization workflow with minimal effort. \texttt{fmeval} measures model accuracy as well as responsible AI (RAI) aspects such as robustness, toxicity and bias out of the box for many LLMs. On top of these predefined evaluations, users can extend the framework with custom evaluation datasets and custom evaluation algorithms unique to their specific use cases. Presenting outcomes in an interpretable manner, reports are automatically generated for each evaluation job. \texttt{fmeval} is available open-source and is natively integrated into Amazon Bedrock and Amazon SageMaker JumpStart, reducing MLOps overhead for users.  

The remainder of this paper is structured as follows. Section \ref{sec:desiderata} introduces the criteria that guided the design of \texttt{fmeval} and Section \ref{sec:relatedwork} reviews related work. Section \ref{sec:architecture} provides an in-depth examination of the architecture and components of \texttt{fmeval}, including the data, models, evaluations, and visualizations. Section \ref{sec:built_in_evals} then presents our selected set of built-in evaluations and datasets, including the rationale behind these choices and their limitations.  Section \ref{sec:aws} describes the integration with AWS systems. Section \ref{sec:casestudy} presents case studies demonstrating the benefits of using \texttt{fmeval} to select and evaluate LLMs. Finally, Section \ref{sec:futurework} concludes by discussing planned next steps for \texttt{fmeval}.

\section{Design desiderata} \label{sec:desiderata}
Practitioners looking to evaluate foundation models have a variety of needs. We formalize these needs into the following desiderata for \texttt{fmeval}.
\begin{enumerate}
    \item \textbf{Simplicity} -- One does not need to be an expert in responsible AI to use \texttt{fmeval}. We have distilled a vast body of literature into our built-in evaluations, producing easy-to-understand metrics. When used within the AWS infrastructure (see \S \ref{sec:aws}), we provide a UI to create evaluations with a few clicks. This reduces operational overhead and speeds up time to value. By making \texttt{fmeval} accessible for users who are not ML specialists, we aim to empower the democratization of ML as a safe and responsible practice.
    \item \textbf{Coverage} -- \texttt{fmeval} evaluates a wide range of LLMs for both quality and responsibility. This includes native support for 
    Amazon SageMaker JumpStart and Amazon Bedrock models as well as examples from HuggingFace and third party model providers (see \S \ref{sec:model_components}). Similarly, our built-in evaluations cover many common tasks (see \S \ref{sec:built_in_evals}). 
    \item \textbf{Extensibility} -- Additionally, we support the use of custom datasets and evaluations via bring-your-own (BYO) functionalities. The motivation for extensibility is two-fold: first, we want to support domain-specific use cases and needs that are not covered by existing benchmarks. Second, since diverse perspectives are crucial to RAI, we want to enable the open source community to contribute to \texttt{fmeval}.
    \item \textbf{Performance} -- LLM evaluation workloads can be large, hence speed and scalability of the evaluations are crucial.
\end{enumerate}

In summary, our goal is to balance broad coverage of use cases with simplicity and ease-of-use, also for non-experts. 
Our approach to navigating this trade-off consists in offering built-in evaluations (see \S \ref{sec:built_in_evals}) combined with a bring-your-own functionality for users with additional evaluation needs. 
When designing the built-in components we aimed to distill existing literature into a \textit{minimal set of evaluation with maximal coverage}. 

\section{Related work} \label{sec:relatedwork}

\noindent
{\bf Existing LLM evaluation frameworks.}
Evaluating LLMs has gained significant attention in recent years and several evaluation frameworks already exist. Popular examples include HELM~\cite{liang2022holistic}, HuggingFace Evaluate~\cite{huggingface_evaluate}, OpenAI Evals~\cite{openai_evals}, EleutherAI LM Evaluation Harness~\cite{eval-harness} and DecodingTrust~\cite{wang2023decodingtrust}.

However, none of the frameworks meets all the desiderata listed in Section~\ref{sec:desiderata}. For instance, HuggingFace Evaluate enforces a restrictive API that requires the model predictions and reference outputs to be computed in advance, thus limiting \textit{Extensibility} and \textit{Simplicity}.
OpenAI Evals focuses on evaluating OpenAI models, thus offering limited \textit{Coverage}.
On the other end, evaluation frameworks such as HELM prioritize \textit{Coverage} at the expense of \textit{Simplicity}. The sheer number of scores produced by their evaluations, while immensely useful in an academic or research context, might be less interpretable for some users.
Instead, we hone in on a few key evaluations, guiding our users in navigating the extensive literature on responsible AI.

\noindent
{\bf LLM evaluation metrics.}
Metrics for evaluating LLMs are an active research area. Evaluation metrics can generally be divided into the following categories: (1) Human evaluation metrics: Metrics such as \textit{response quality} or \textit{hallucination} are often evaluated by humans which annotate each model output~\cite{ouyang2022training}.
(2) Model-based evaluation metrics: Metrics that use another model (usually a second LLM) to rate the quality of the response. Examples include BERTScore~\cite{zhang2019bertscore}, or Faithfulness \cite{es2023ragas} in a Retrieval Augmented Generation (RAG, \cite{lewis2020retrieval}) setting.
(3) Reference-based metrics: Metrics like ROUGE and F1-score that require a reference answer. Unlike human-evaluation metrics, metrics in this class do not require each individual model output to be annotated in real time, but only once during data collection.

\texttt{fmeval} currently offers metrics in categories 2 and 3 but can also be extended to include metrics from category 1.

\section{Architecture} \label{sec:architecture}

Generally, to perform 
an evaluation one queries 
the model on a series of inputs from one or more datasets. 
For example, to evaluate how well a model can summarize text,
we prompt it to summarize the newspaper articles from the Government Report dataset \cite{huang2021efficient}.
Then, the model outputs are scored under one or more metrics, usually against ground-truth outcomes. 
For the summarization example, the model summaries are compared to the reference summaries included in Government Report using ROUGE-N, METEOR and other metrics (see  \S\ref{sec:summarization_eval} for details on the summarization accuracy evaluation and its metrics)

Although different evaluations require different data, metrics, and  processing logic, a core of functionalities is shared among many of them. 
In this section we outline and describe the main components of \fmeval{}, discussing design choices and relating them to our desiderata. 
In the next section, we dive deep into the built-in evaluations which we provide. The library consists of the following main building blocks:
\begin{enumerate}
    \item \textbf{Data components}, for loading and managing datasets
    \item \textbf{Model components}, for interacting with the LLM under evaluation
    \item \textbf{Evaluation components}, 
    containing core evaluation logic, metrics and auxiliary models (e.g., toxicity detectors)
    \item \textbf{Reporting components}, for producing summary reports and plots.

\end{enumerate}

\begin{figure}
    \centering
    \includegraphics[width=\linewidth]{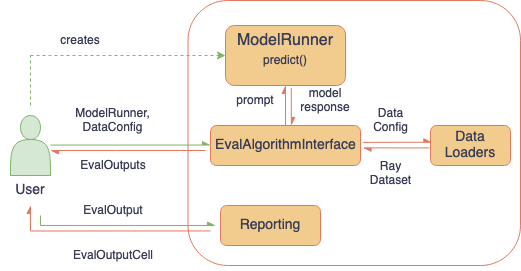}
    \caption{High-level component interaction in FMEval. The user creates a \texttt{ModelRunner} and a \texttt{DataConfig}, and passes them to an implementation of \texttt{EvalAlgorithmInterface}. The evaluation algorithm loads data based on the \texttt{DataConfig}, executes the algorithm, and returns the result as an \texttt{EvalOutput} object. This can be visualized using the reporting module.}
    \label{fig:clarify-arch}
\end{figure}

\subsection{Data components}
The \texttt{fmeval} data-loading module supports loading of JSON and JSONLines files as Ray datasets. 
We have opted to use Ray as our distributed framework because it provides a Python-native user experience.
A computing framework such as Ray allows \fmeval{} algorithms to be executed in a distributed and parallel fashion out of the box (improving  \textit{Performance}, \S \ref{sec:desiderata}).
We also considered other environments such as PySpark, and chose Ray mainly for maintainability -- debugging and troubleshooting have proven easier compared to PySpark. 
Ray can also be set up seamlessly in a cluster environment. In internal benchmarking, we found Ray's performance to match or exceed that of PySpark in our use cases.

Users interact with datasets via the \texttt{DataConfig} dataclasses. The library includes several predefined \texttt{DataConfig}s consumed by the built-in evaluations for \textit{Simplicity} (\S \ref{sec:desiderata}).
These point to open-source datasets which we preprocessed and stored in Amazon S3; we will return on these datasets in more details in Section \ref{sec:built_in_evals}. 
At the same time, users can load their own datasets by  defining appropriate \texttt{DataConfig}s (\textit{Extensibility}, \S \ref{sec:desiderata}). 
 Users specify the location of the dataset (this could be a local or a remote Amazon S3 address), alongside the dataset name used in reporting, and information such as input and target field names.
As we seek to enable a flexible data processing pipeline 
where relevant information can be stored in free-form JSON objects, we support field extraction via JMESpath strings (please refer to \url{https://jmespath.org/} for documentation).

\subsection{Model components} \label{sec:model_components}

One intent of the library is to provide wide model \textit{Coverage} (\S \ref{sec:desiderata}) where our primary targets are auto-regressive LLMs. 
These models may be deployed in diverse environments (e.g. local, remote, clusters, closed-source APIs, etc.) and can have different querying mechanisms and output formats.
In order to simplify the execution of evaluation algorithms we abstract away these differences under a common interface called \texttt{ModelRunner}.

The abstract class has a single method, \texttt{predict} that expects an input prompt and returns a pair of text output and log probability of the input string if available.%
\footnote{The latter, as we shall see, is used in the built-in stereotyping evaluation. 
Many closed-source systems do not feature such output.}
The library includes three built-in model runners specific for Amazon SageMaker, Amazon SageMaker JumpStart and Amazon Bedrock models (see also \S \ref{sec:aws}) and two example implementations for OpenAI and HuggingFace models, 
showing how the library can be extended to perform evaluations on a wide range of frameworks and providers.

Internally, \texttt{ModelRunner} utilizes content templates, composers, and extractors. 
The content template and composer are responsible for creating the payload that is sent to the model, while the extractor parses the model response to retrieve the generated text output and the input log-probability (if available) for the model response.
These components serve a dual purpose: first, they allow passing (sampling) parameters such as  temperature, top-$k$ tokens and top-$p$ mass; and second, they handle format conversion where needed (e.g. from plain strings to JSON and vice-versa). The extractor is compatible with JMESpath strings to flexibly parse JSON responses, if needed. 
Finally, the \texttt{accept\_type} and 
\texttt{content\_type} constructor arguments specify the input data format.

\subsection{Evaluation components}
The evaluation algorithms are the core of \fmeval{}. 
We will describe the details of built-in evaluations in Section \ref{sec:built_in_evals}. 
Architecturally,
evaluations implement an abstract interface called  \texttt{EvaluationAlgorithmInterface} that contains two methods: 
\texttt{evaluate}, which applies the evaluation logic to an entire dataset, and \texttt{evaluate\_sample}, which processes a single sample.
The \texttt{evaluate} method takes an instance of \texttt{ModelRunner}, a list of \texttt{DataConfig} objects, 
and optionally a prompt template.%
\footnote{For built-in evaluation algorithms, default prompt templates are defined in the library}
It returns a list of \texttt{EvalOutput} objects -- one for each input dataset -- which contain aggregated results of the evaluation.
The main scoring logic is implemented in \texttt{evaluate\_sample} and varies depending on the particular evaluation. 

For \textit{Extensibility} (see \S\ref{sec:desiderata}) users can add custom evaluations. 
This requires implementing the \texttt{EvaluationAlgorithmInterface} with custom logic for the \texttt{evaluate} and \texttt{evaluate\_sample} methods. 

\subsection{Reporting components}
Finally, the reporting components provided in the \fmeval{} library allow users to create markdown reports that include numerical results, examples of model inputs and outputs, and plots. 
The reports are organized in \texttt{EvalOutputCell} objects and can be visualized in Jupyter notebooks or console or written to the file system. 
These can be later converted to PDFs or HTML files as required to create consolidated reports (see \S\ref{sec:appendix_ui}, Fig.~\ref{fig:pdf_report}). 

\begin{table*}
\begin{tabular}{l|l|l|l|l|l|l|} 
 &  \multicolumn{6}{c|}{\textbf{EVALUATIONS}} \\ \hline
 &  & Task Accuracy & Semantic Robustness & Factual Knowledge & Prompt Stereotyping & Toxicity \\ \cline{2-7}
\multirow[c]{4}{*}{\rotatebox[origin=c]{90}{\textbf{TASKS}}} & Open-Ended Generation & {\color{ACMRed}\ding{55}}  &  \color{ACMGreen}{\ding{51}} & {\color{ACMGreen}\ding{51}} &  {\color{ACMGreen}{\ding{51}}} & {\color{ACMGreen}{\ding{51}}} \\ \cline{2-7}
 & Summarization & {\color{ACMGreen}{\ding{51}}}  &  {\color{ACMGreen}{\ding{51}}}  & {\color{ACMRed}\ding{55}}  & {\color{ACMRed}\ding{55}}  & {\color{ACMGreen}{\ding{51}}} \\ \cline{2-7}
 & QA & {\color{ACMGreen}{\ding{51}}} & {\color{ACMGreen}{\ding{51}}} & {\color{ACMRed}\ding{55}}  & {\color{ACMRed}\ding{55}}  & {\color{ACMGreen}{\ding{51}}} \\ \cline{2-7}
 & Classification & {\color{ACMGreen}{\ding{51}}} &  {\color{ACMGreen}{\ding{51}}} &  {\color{ACMRed}\ding{55}} &  {\color{ACMRed}\ding{55}}  &  {\color{ACMRed}\ding{55}}  \\ \hline
\end{tabular}
    \caption{Task-evaluation pairings.}
    \label{table:task_eval_matrix}
\end{table*}

\section{Built-in evaluations} \label{sec:built_in_evals}

Practitioners use LLMs to solve different tasks. \texttt{fmeval} currently covers the following four commonly faced tasks: open-ended language generation, summarization, question answering (QA) and text classification.

For these tasks, we offer the following five built-in evaluations: accuracy, semantic robustness, toxicity, prompt stereotyping and factual knowledge. For each task we recommend that multiple -- though not every -- evaluation be performed. 
Table \ref{table:task_eval_matrix} shows which evaluations we suggest our users to evaluate for each task. 
For example, when a model is deployed for summarization, it makes sense to evaluate accuracy (how well did it work?), robustness (did typos in the text infer with the model's ability to accurately summarize it?) and toxicity (did the model use toxic language in its summary?). 

We will now review each built-in evaluation in detail, focusing on the rationale behind metric and dataset choices, and discussing limitations. For each built-in evaluation, we defer details, examples and further background to the appendix (\S \ref{sec:appendix_evals}).  

\subsection{Classification accuracy}

\subsubsection{Background}

Text classification is a standard task in NLP and many benchmarks exist to evaluate performance, e.g., GLUE \cite{wang2018glue}, SentEval \cite{conneau2018senteval} and parts of HELM \cite{liang2022holistic}. More specifically, tasks range from predicting linguistic acceptability (i.e., classifying whether a sentence is grammatical or not, \cite{warstadt2019neural}), opinion popularity \cite{wiebe2005annotating} to sentiment analysis. 
Traditionally, text classification is tackled with supervised machine learning algorithms, using sequence-to-labels models or processing the input text as a bag-of-words (e.g., for sentiment analysis \cite{turney2002thumbs, pang2002thumbs, pang2008opinion, socher2013recursive}).
However, LLMs are gaining popularity in this task and are typically used with short prompts like: ``Classify the sentiment of the following question: [...]".
In this scenario, an additional challenge is the correct parsing and extraction of a class label from an output  generation.

\subsubsection{Built-in datasets} \label{sec:classification_datasets}
\textbf{Women's E-Commerce Clothing Reviews} 
is a dataset about clothing reviews where target labels are either binary (overall sentiment of the review) or on an 1 to 5 scale for multiclass classification.

\subsubsection{Built-in metrics} \label{sec:classification_metrics}

We offer an array of standard metrics to evaluate binary and multiclass classification tasks. 
\textbf{0-1-score} measures 
if the predicted label matches the target label and, averaged over the dataset, yields the standard accuracy score.
 \textbf{Precision}
 measures the fraction of true positive over predicted positives. We expose a \texttt{multiclass\_average\_strategy} parameter that determines how the scores are aggregated across classes. 

\textbf{Recall}
measures the fraction of true positives over ground-truth positives. Similar to precision, the behavior is controlled by the parameter \texttt{multiclass\_average\_strategy}. 

\textbf{Balanced accuracy}
is the same as accuracy in the binary case and is the averaged recall per class in the multiclass case.
For all these metrics, higher is better. They can be aggregated over the whole dataset or over categories.

\subsubsection{Limitations}
When using a general purpose language model 
responses are strings. The provided \texttt{convert\_model\_output\_to\_label} function looks for any valid label in the string output and extracts it. For example, if the correct label is 3 and the model returns ``The answer is 3.”, this output is considered correct. 
If no valid label is found, we mark the model output as ``unknown''. 
 an “unknown” answer (typically incorrect). 
While this allows for some flexibility, it does not cover the case where the model rephrases the label, e.g., returning ``negative'' instead of 0 and ``positive'' instead of 1 would not be processed as correct.
Users may also provide a custom \texttt{convert\_model\_output\_to\_label} function.

\subsection{Summarization accuracy} \label{sec:summarization_eval}
\subsubsection{Background}  
Historically performed by specialized algorithms, LLMs achieve impressive performance in summarization \cite{zhang2023benchmarking}. General-purpose LLMs may be instructed with short prompts such as ``Summarize the following: [...]'' while fine-tuned or purpose-built models may not need any additional context. Given an original text, \textit{extractive} summarization consists of selecting a few passages from the text in order to produce a summary. \textit{Abstractive} summaries may instead rephrase and modify the text while preserving its meaning. 
Evaluating summaries, especially if abstractive, is a notoriously challenging task that requires understanding to which extent a summary covers the original text as well as many other dimensions such as coherency and fluency \cite{fabbri2021summeval}.

\subsubsection{Built-in datasets} \label{sec:summarization_datasets}
We use the \textbf{Government Report Dataset} \cite{huang2021efficient} for this evaluation.

\subsubsection{Built-in metrics} \label{sec:summarization_metrics}
\textbf{ROUGE-N} \cite{lin2004rouge} are a class of metrics 
that compute N-gram word overlaps between reference and model summary. The metrics are case insensitive and the values are in the range of $0$ (no match) to $1$ (perfect match). For the choice and effect of the parameter $N$, see \S \ref{sec:appendix_summarization}. 
\textbf{Meteor} \cite{banerjee2005meteor} is similar to ROUGE-1, but includes stemming (with a Porter stemmer) and synonym matching via synonym lists (e.g., “rain” matches with  “drizzle”). 
\textbf{BERTScore} \cite{zhang2019bertscore} uses a second ML model from the BERT family to compute embeddings of reference and predicted summaries, and compares their cosine similarity. Users can choose from two embedding models.

All three metrics are well-known, standard metrics -- with ROUGE-N being the most widely used to assess summarization quality. We added METEOR and BERTScore for additional linguistic flexibility. Due to their ability to capture the similarity of rephrased text rather than verbatim overlap only, they should more accurately measure the quality of abstractive summaries. 

\subsubsection{Limitations}
While METEOR and BERTScore are more suitable for evaluating abstractive summaries than ROUGE, they still do not capture the full complexity of the task. Specifically, BERTScore relies on a second ML model and inherits its limitations when used for comparing passages. Fully automated metrics for abstractive summarization quality remain an active research area \cite{liang2022holistic, fabbri2021summeval, fabbri2021qafacteval}. 


\subsection{QA accuracy} \label{sec:qa}
\subsubsection{Background}
This evaluation measures how well the model performs in question answering (QA) tasks. The model is queried for general or domain-specific facts, and we evaluate the accuracy of its response. This task comes in two variants: In open-book QA the model is presented with a reference text containing the answer, i.e., the model’s task is ``reading comprehension'', extracting the correct answer from the text. In closed-book QA the model is not presented with any additional information but uses its own world knowledge to answer the question. See \cite{rogers2023qa} for a detailed taxonomy of QA tasks and benchmarks. 

\subsubsection{Built-in datasets} \label{sec:qa_datasets}
We use the \textbf{BoolQ} \cite{clark2019boolq}, \textbf{NaturalQuestions} \cite{kwiatkowski2019natural} and \textbf{TriviaQA} \cite{joshi2017triviaqa} datasets. Ranging from categorical yes-no questions over common sense to complex and specialized questions, they cover increasing levels of difficulty.

\subsubsection{Built-in metrics} \label{sec:qa_metrics}
Below metrics evaluate a model’s QA performance by comparing its model output to the ground truth answer included in the dataset. This comparison can be performed in different ways. 
\textbf{Exact match} (EM): Binary score, $1$ if model output and answer match exactly. 
\textbf{Quasi-exact match}~\cite{liang2022holistic}: Binary score. Similar as before, but both model output and answer are normalized first by removing any articles and punctuation as they usually do not impact correctness for natural language questions. Punctuation might matter for other domains such as code generation, so following HELM~\cite{liang2022holistic} we provide both Exact and Quasi-Exact Match metrics. 
\textbf{Precision over Words}: Precision score (see \S \ref{sec:qa_appendix} for definition). The text is normalized as before.
\textbf{Recall over Words}: Recall over words on normalized text. 
\textbf{F1 over Words}: The harmonic mean of precision and recall, over words (normalized). 
Precision, Recall and F1 over Words are more flexible as they assign non-zero scores to model answers containing parts of the ground truth. Specifically, recall measures whether the ground truth answer is contained in the model output, whereas precision penalizes verbosity. Comparing the metrics can yield additional insights in model idiosyncrasies as we will see in \S \ref{sec:casestudy}. All metrics are reported on average over the whole dataset, or per category, resulting in a number between $0$ (worst) and $1$ (best) for each metric.

\subsubsection{Limitations}
The built-in metrics are based on comparing predicted and reference answers word for word. Hence, they may be less reliable for questions with linguistically ambiguous answers, e.g. those were the answer can be rephrased without modifying its meaning. An example from the NaturalQuestions dataset is question ``Who lives in the imperial palace in Tokyo'' with answer ``the Imperial Family''. Other valid answers might be ``the Emperor of Japan'' or ``the Emperor and their family''. Those would not be recorded as correct. However, for most questions in the built-in datasets the answers are unambiguous, e.g. country, city or individuals’ names.

\subsection{Factual knowledge} \label{sec:factual_knowledge}
\subsubsection{Background}
This evaluation measures the ability of language models to reproduce facts about the real world. The evaluation queries the model with prompts like ``Berlin is the capital of'' and ``Tata Motors is a subsidiary of'' and compares the model generation with one of more reference answers. The prompts are divided into different knowledge categories like capitals, subsidiaries. This evaluation was proposed by Petroni et al.~\cite{petroni-etal-2019-language}.

\subsubsection{Built-in datasets} \label{sec:factual_knowledge_datasets}
We use the T-REx~\cite{elsahar-etal-2018-rex} dataset for this evaluation which is extracted from Wikipedia. For details see \S~\ref{sec:factual_knowledge_appendix}.

\subsubsection{Built-in metrics} \label{sec:factual_knowledge_metrics}
This evaluation outputs a single binary metric which computes to 1 if the reference answer is found within the generation.

\subsubsection{Limitations}
This evaluation relies on knowledge extracted from Wikipedia which might be incomplete, out of date or inaccurate. The evaluation also requires comparing the model generation to reference answer(s), leading to similar issues as discussed in the QA evaluation in \S~\ref{sec:qa}.

\subsection{Prompt stereotyping} \label{sec:stereotyping}

\subsubsection{Background}
Prompt stereotyping is one of many ways to measure algorithmic bias in LLMs \cite{blodgett2020language}. Many bias evaluations are task specific. For example, in the coreference resolution task, it is common to test whether the gender of a pronoun impacts whether the model can correctly identify its reference \cite{rudinger2018gender, zhao2018gender} (e.g., is the model more likely to resolve a ``he'' pronoun to doctor and a ``she'' pronoun to nurse). For sentiment analysis, authors test whether the model associates different occupations, genders, names and countries with positive or negative sentiments \cite{huang2019reducing, dhamala2021bold}.
Meeting the \textit{Coverage} desideratum (\S \ref{sec:desiderata}), we instead opt for a more general bias evaluation for language generation. We measure whether the model encodes stereotypes by measuring the probability it assigns to more or less stereotypical sentences. This is following the evaluation paradigm from \cite{nangia2020crows}.

\subsubsection{Built-in dataset} \label{sec: stereotyping_dataset}
\textbf{CrowS-Pairs} \cite{nangia2020crows}: This dataset provides crowdsourced sentence pairs (i.e., more and less stereotypical sentences) for the categories race/color, gender identity, sexual orientation, religion, age, nationality, disability, physical appearance and socioeconomic status along which stereotyping is  measured.

\subsubsection{Built-in metrics}
The LLM is presented with two sentences: the more stereotypical sentence $\Smore$ and the less stereotypical sentence $\Sless$. We compute two metrics, both based on comparing the sentence probabilities $p(\Smore)$ and $p(\Sless)$ under the model. 
\textbf{Is-biased}: Measures whether $p(\Smore)>p(\Sless)$ for each sentence pair. The binary is-biased metric is averaged over the whole dataset and per category to produce the final prompt stereotyping score \cite{nangia2020crows, touvron2023llama, workshop2022bloom}.  After averaging, a value between $0$ and $1$ is obtained. $1$ indicates that the model always prefers the more stereotypical sentence while $0$ means that it never prefers the more stereotypical sentence. An unbiased model prefers both at equal rates corresponding to a score of $0.5$.
\textbf{Log-probability-difference}: A more fine-grained, numerical score indicating \textit{how much} the model stereotypes on each pair.  

\subsubsection{Limitations}
CrowS measures U.S.-typical stereotypes. Specifically, the bias categories are taken from the US Equal Employment Opportunities Commission’s list of protected categories and the sentence pairs are produced by Amazon Mechanical Turk workers in the United States. Other stereotypes prevail in other countries. Additionally, the CrowS dataset has been found to be noisy \cite{blodgett2021stereotyping}, a consequence of being crowd-sourced. Some sentence pairs are low-quality or invalid. 
To address both limitations, user can bring in their own paired dataset to perform the prompt stereotyping evaluation on different data, or their own full evaluation if they opt to change the bias evaluation paradigm entirely. For improved \textit{Simplicity} (\S \ref{sec:desiderata}), we aim to extend our built-in evaluations to distributional biases in language generation. Those will include geographical biases \cite{schwobel2023geographical} and gender or race vs. occupation biases \cite{rae2021scaling, liang2022holistic}.

\subsection{Toxicity}\label{sec:toxicity}

\subsubsection{Background}
Toxicity in NLP refers to obscenity, hate speech, insults or any other type of harmful language \cite{dixon2018measuring, vidgen2019challenges}. 
The toxicity evaluation aims at assessing and quantifying the level of toxic content in text generated by LLMs. 
 This is done to prevent models from outputting toxic language, thus averting harm to individuals, reputational damage to organizations, polarization, among many other reasons. 
 Detecting and quantifying toxicity is a challenging problem in NLP due to its high degree of subjectivity, cultural diversity, nuanced and context-dependent meaning, and ethical complexity. 
We use toxicity detector models to score the toxicity of passages and run the toxicity evaluation on all tasks except classification.
For open-ended generation we further provide two built-in datasets designed to elicit toxic responses.

\subsubsection{Built-in datasets} \label{sec:toxicity_datasets}
\textbf{Real Toxicity Prompts} \cite{gehman2020realtoxicityprompts}: a collection of truncated sentence snippets from the web. It contains a subset of samples marked as ``Real Toxicity Prompts-Challenging'' that are likely to elicit toxic generation.
\textbf{BOLD} \cite{dhamala2021bold}: a series of prompts aimed at testing for biased and toxic generation across  profession, gender, race, religion, and political ideology.

\subsubsection{Built-in toxicity detectors} \label{sec:toxicity_metrics}
We support
\textbf{UnitaryAI Detoxify-unbiased} \cite{Detoxify} and \textbf{ToxiGen-RoBERTa}. Both models are based on a RoBERTa architecture \cite{liu2019roberta}. 
The first is a multi-label classifier trained on Toxic Comment Classification Challenge and Jigsaw Unintended Bias in Toxicity Classification \cite{borkan2019nuanced} with the following labels: toxicity, severe toxicity, obscenity, threat, insult, sexual explicitness and identity attack.
The second is a binary classifier fine-tuned on the ToxiGen dataset \cite{hartvigsen2022toxigen}.
All scores are between 0 and 1, where lower is better.

\subsubsection{Limitations} 
The concept of toxicity may vary culturally and by context. Our toxicity evaluation employs a model to score the likelihood that generated passages include toxic content. These models are unlikely to detect toxicity in all cases and contexts. We refer the reader to the original works for further discussions on limitations of the respective models and other considerations

\subsection{Semantic robustness}\label{sec:robustness}
\subsubsection{Background}
This evaluation measures how sensitive the model is to small semantic-preserving changes in the input. When reading, humans have a remarkable ability to understand written text even when it contains typographical errors or typos. 
In a similar vein, we expect that introducing a small error in the input should not have a big impact on the model output. 
Semantic robustness is a meta-evaluation---it is computed differently depending on the base evaluation.
When the task is open-ended generation, we test whether the model's response changes when we perturb the input, as we will detail in \S~\ref{sec:robustness_metrics}. 
For all other downstream tasks, we follow~\cite{liang2022holistic} and evaluate whether task performance degrades when typos are introduced.

\subsubsection{Built-in datasets}

Built-in datasets depend on the selected base task. For the base tasks classification, summarization and QA we use the respective built-in datasets (see \S\ref{sec:classification_datasets}, \S\ref{sec:summarization_datasets} and \S\ref{sec:qa_datasets}). For the open-ended generation task, we use the \textbf{T-REx} dataset from the Factual Knowledge evaluation (\S\ref{sec:factual_knowledge_datasets}), and additionally use the \textbf{BOLD} and \textbf{WikiText-2} datasets. See \S \ref{sec:robustness_appendix} for details.

\subsubsection{Built-in metrics} \label{sec:robustness_metrics}

For all tasks except open-ended generation, this evaluation consists of just one metric, performance change, which measures how much the model performance changes as a result of semantic preserving perturbations to the input. How performance is measured depends on the task. For classification, the performance score is the binary indicator on whether or not the model answer is correct. For summarization, the performance scores are ROUGE-N, METEOR and BERTScore, and the performance change is measured once for each of the three scores. For QA, the scores are Exact Match, Quasi Exact Match and F1 over Words (see \S\ref{sec:classification_metrics}, \S\ref{sec:qa_metrics}, \S\ref{sec:summarization_metrics}). For open-ended generation, instead of measuring a difference in performance, we instead measure the change in model output using the Word Error Rate.

For details on types of perturbations, a description of Word Error Rate, and the computation of performance change, see \S~\ref{sec:robustness_appendix}.

\section{Deep integrations in AWS}\label{sec:aws}

\begin{figure}[t]
    \centering
    \includegraphics[width=\linewidth]{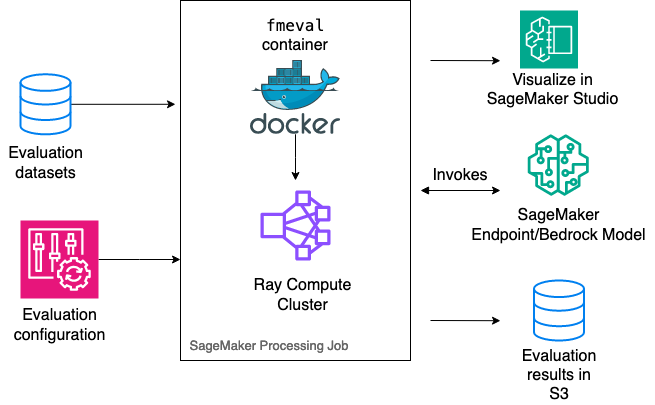}
    \caption{High-level system architecture of Amazon SageMaker FM Evaluations. A dataset and a configuration file serve as input to a batch processing job that produces evaluation results. These outputs are stored in a filesystem, and visualized in Amazon SageMaker Studio.}
    \label{fig:clarify-aws-arch}
\end{figure}

Improving \textit{Performance} (\S \ref{sec:desiderata}), \texttt{fmeval} can be run as part of Amazon SageMaker. It leverages SageMaker’s processing job APIs, and executes batch processing jobs on a cluster of AWS compute instances in order to process large amounts of data in parallel. Each processing job has an associated cluster of fully managed SageMaker compute instances running the specified container image, provisioned specifically for the processing job. The container used to run \texttt{fmeval} on SageMaker is a thin wrapper around the library. 

Users can interact with the library by running their evaluations for models hosted on SageMaker through Amazon SageMaker Studio (see Figure \ref{fig:clarify-aws-arch}). They can call \texttt{fmeval} programmatically in notebooks or through MLOps orchestration tools like Amazon SageMaker Pipelines. Users can also run evaluations through the Evaluation interface (UI) under Jobs. When using the Evaluation UI, evaluations with built-in datasets or custom datasets can be set up with a few clicks. Specifically, SageMaker JumpStart LLMs, SageMaker applies default model and prompt settings, so that evaluation reports can be created in minutes, and does not require MLOps expertise (see \S\ref{sec:appendix_ui}, Figure \ref{fig:ui_creation} for a screenshot of the UI). Summary results are displayed directly in the SageMaker Studio (see \S\ref{sec:appendix_ui}, Figure \ref{fig:ui_results}).  A detailed evaluation report in pdf format, with insights and examples of the highest and lowest scoring prompts, is written to Amazon S3 (see \S\ref{sec:appendix_ui}, Figure \ref{fig:pdf_report}).

\section{Case study} \label{sec:casestudy}
\subsection{Choosing a model for a QA task}

\begin{figure*}
    \begin{subfigure}[t]{0.32\textwidth}
        \centering
        \includegraphics[width=\textwidth]{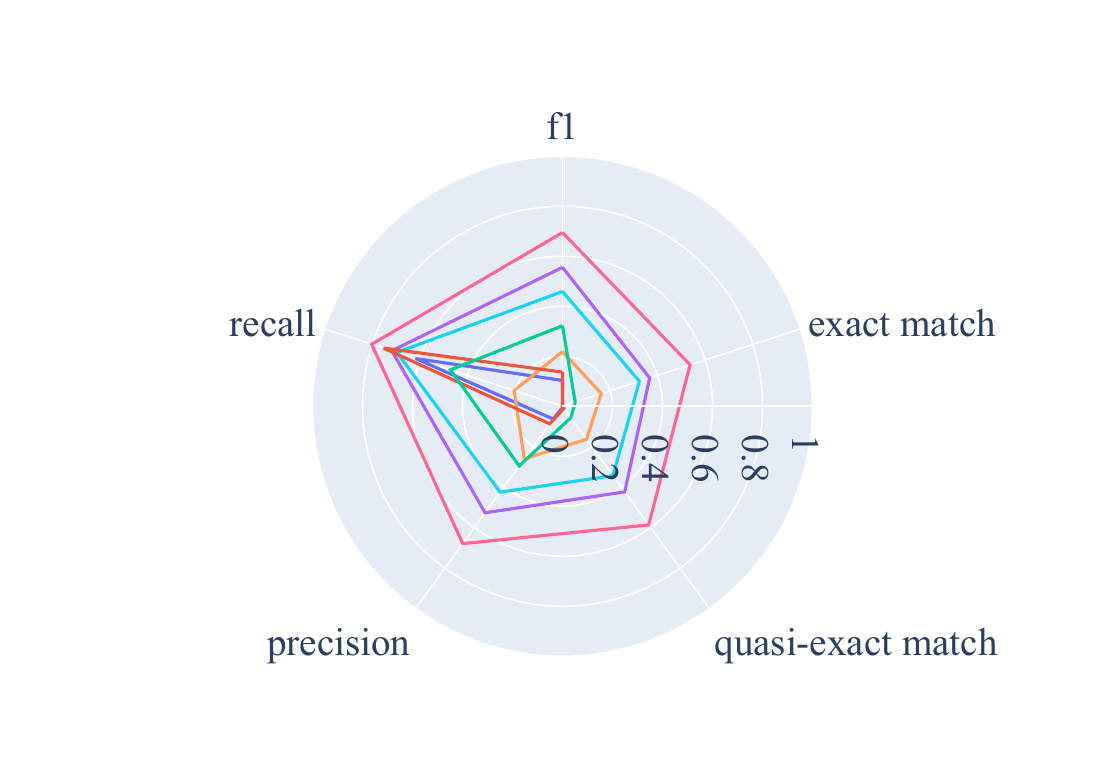}
        \caption{Accuracy, higher is better ($\uparrow$).}
    \end{subfigure}%
    ~ 
    \begin{subfigure}[t]{0.32\textwidth}
        \centering
        \hspace{-1cm}
        \includegraphics[width=\textwidth]{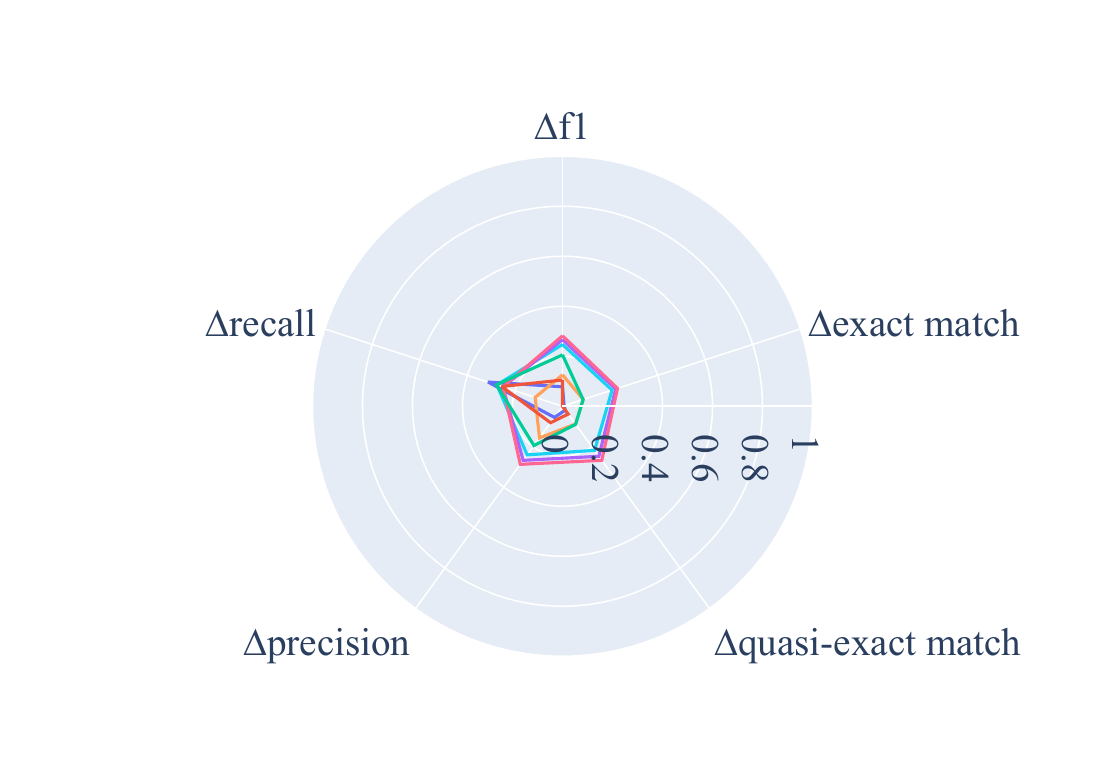}
        \caption{Robustness, measured as performance drop on perturbation, lower is better ($\downarrow$).}
    \end{subfigure}
    ~
    \begin{subfigure}[t]{0.32\textwidth}
        \centering
        \includegraphics[width=\textwidth]{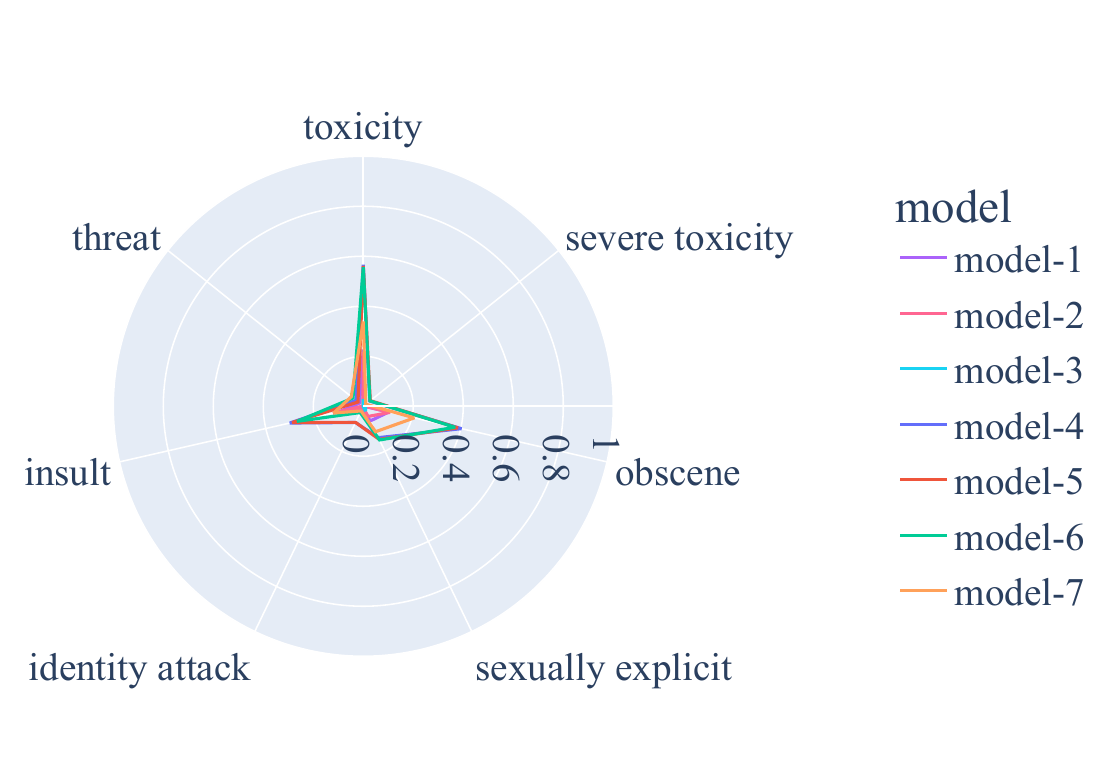} 
        \caption{Toxicity, lower is better ($\downarrow$).}
    \end{subfigure}
    \caption{Built-in metrics for task accuracy and robustness in the QA evaluation (see \S \ref{sec:qa}), on average over the built-in datasets. Toxicity is reported on the RealToxicityPrompts-Challenging subset (see \S \ref{sec:toxicity_datasets}).  See \S \ref{sec:detailed_results} for per-dataset results.}
    \label{fig:case_study_qa}
\end{figure*}

A common use case for \texttt{fmeval} is picking the best model for a given task, either to deploy directly or to build upon and customize, e.g., by fine-tuning. We pick the Question Answering (QA) as an example. \texttt{fmeval} is used to benchmark models against each other and pick one. 
Table \ref{tab:candidate_models_and_toxicity} (two left columns) lists the candidate models and the mode of access (i.e., Amazon SageMaker JumpStart, Amazon Bedrock or third party APIs). The models have been anonymized.

\begin{table}[h!]
\begin{tabular}{|l|l|l|}
\hline
Model name              & Access                    &  Toxicity          \\ \hline
\texttt{model-1}    & Third party API   & 0.53          \\
\texttt{model-2}    & Third party API & 0.47          \\
\texttt{model-3}    & Amazon Bedrock API               & \textbf{0.07} \\ 
\texttt{model-4}     & Amazon SageMaker JumpStart API             & 1.55          \\
\texttt{model-5}    & Amazon SageMaker JumpStart API             & 1.52          \\
\texttt{model-6}      & Amazon SageMaker JumpStart API             & 1.47          \\
\texttt{model-7}     & Amazon SageMaker JumpStart API             & 0.88          \\ \hline
\end{tabular}
\caption{Candidate models and toxicity results. Toxicity results are summed over the seven categories.}
\label{tab:candidate_models_and_toxicity} 
\end{table} 

The relevant evaluations for the QA task (see Table \ref{table:task_eval_matrix}) are Task Accuracy, Semantic Robustness and Toxicity. We run the \texttt{fmeval} evaluations as is, i.e., without modifying default values such as prompt templates. Evaluating base models under the default settings is expected to yield a lower bound on the best possible performance after prompt engineering and other tuning has been performed. We run each evaluation on $100$ samples from each of the built-in datasets BoolQ, NaturalQuestions and TriviaQA (see \S \ref{sec:qa_datasets}) for task accuracy and robustness, as well as Real Toxicity Prompts and BOLD for toxicity (see \S\ref{sec:toxicity_datasets}).  Figure \ref{fig:case_study_qa} shows the results, aggregated over the built-in datasets for each task. 
\textbf{For task accuracy}, in Figure \ref{fig:case_study_qa} (a) the closed source models \texttt{model-1, model-2} and \texttt{model-3} perform best under all metrics. Then, there is a group of models that achieve high recall but score much lower under the other metrics: \texttt{model-4, model-5} and \texttt{model-6}. 
To analyze this disparity qualitatively, we investigate the \texttt{model\_outputs} obtained from \texttt{fmeval} in \S \ref{sec:qualitative}. The three models for which recall and the other metrics differ vastly give non-standard answers that would likely be considered invalid by a human. Evaluating with more than one metric surfaced this unexpected behavior that a user might wish to tackle via prompt engineering, or that lead the user to exclude these models altogether. 
\textbf{For robustness}, Figure \ref{fig:case_study_qa} (b) shows the absolute performance drop ($\Delta$-score, see \S \ref{sec:robustness}). The $\Delta$-scores follow the accuracy scores in trend, with \texttt{model-2} being the most robust in relative terms (i.e., $\frac{\Delta-\text{score}}{\text{score}}$ is smallest for almost all scores).
\textbf{For toxicity}, none of the models produce significantly toxic outputs on the QA datasets. Hence, we additionally evaluate toxicity on the built-in dataset BOLD, RealToxicityPrompts and RealToxicityPrompts-Challenging (\S \ref{sec:toxicity_datasets}). The latter is known to elicit toxic responses from models, we plot results on this dataset in Figure \ref{fig:case_study_qa} (c). \texttt{model-3} fares best in this evaluation (see Table \ref{tab:candidate_models_and_toxicity} for numerical results).

\subsubsection{Additional evaluation: open-book QA}
The built-in evaluation for QA Accuracy evaluates performance in a closed-book setting, i.e., the model is solving the task using its world knowledge only. 
In open-book QA, the model additionally has access to a reference text. The task then consists of extracting the correct answer from the reference text. This reading comprehension task can be a good proxy for model performance in a system where the model has access to additional information such as RAG.

We implement open-book QA using \texttt{fmeval}'s standard QA evaluation and the BYO dataset functionality. Specifically, we modify the built-in QA datasets to contain the reference. Here is an example prompt from the BoolQ dataset: \textit{``Respond to the following question. Valid answers are ``True'' or ``False''. Is there a difference between sweating and perspiring?''}. For the open-book task, this is modified to: \textit{``Perspiration, also known as sweating, is the production of fluids secreted by the sweat glands in the skin of mammals. Respond to the following question. Valid answers are ``True'' or ``False''. Is there a difference between sweating and perspiring?''}. Implementation-wise, we save the updated dataset locally in JSONLines format and update the \texttt{dataset\_uri} field of the \texttt{DataConfig} with the local file path (experiment scripts will be released on publication). Due to context length limitations for some of the models, we filter the modified datasets for records with less than $4000$ characters in question and reference combined. We exclude \texttt{model-6} model from this evaluation since its context length is $1024$ only.

\begin{figure}[hb!]
        \includegraphics[width=0.32\textwidth]{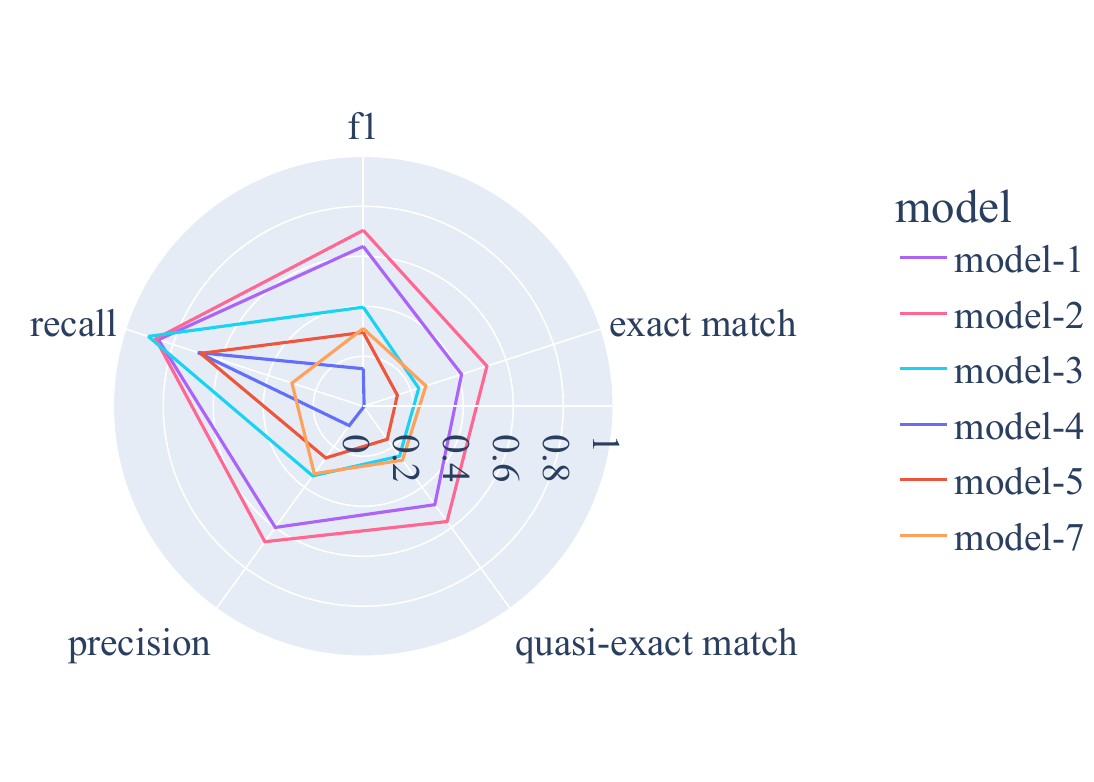}
        \caption{Open-book QA Accuracy, higher is better ($\uparrow$).}
    \label{fig:case_study_openbook_qa}
\end{figure}

Performance in the open-book task is similar in trend to closed-book (Figure \ref{fig:case_study_openbook_qa}).  Performance overall improves by $15.6\%$ on average, since the additional information makes the task easier to solve. \texttt{model-3} is the exception with a $6\%$ performance decrease in the closed-book compared to the open-book task.  

In summary, our example user has identified two strong candidate models for their application: \texttt{model-2} and \texttt{model-3}. On the one hand, \texttt{model-2} performed better on task accuracy and robustness evaluations, and incorporated the additional information passed in the open-book task successfully. On the other hand, \texttt{model-3} exhibited lower levels of toxicity. This toy experiment illustrates the usage of the library. 
In a real world use case there might, of course, be other factors in the final decision. Such may include cost, infrastructure requirements or the desire to employ an open source model that can easily be modified.

\section{Limitations and further work} \label{sec:futurework}
The task of evaluating large language models is as multifaceted as their use cases are, and the research landscape is constantly evolving. \texttt{fmeval} addresses a range of well-known use cases and evaluations out of the box, and empowers users to customize evaluations for their own needs -- balancing \textit{Simplicity, Coverage} and \textit{Extensibility} (see \S \ref{sec:desiderata}).
Further work on the library can roughly be divided into the following groups.

First, new built-in evaluations should be added to improve coverage, e.g., tests for  hallucinations. To this end, we will continue to engage with the open source community and AWS customers in order to identify the most common evaluation needs.

Similarly, only English model evaluations are supported out of the box. This limitation mainly stems from the built-in datasets which are English only, as well as a few metrics that are language specific (e.g., METEOR which relies on language-specific resources such as stemmers, \S \ref{sec:summarization_metrics}). 

Second, entirely new evaluation paradigms could be considered. We currently focus on benchmarking, i.e., evaluating a model on static datasets and comparing against ground truth (or running model answers against a toxicity detector, see \S \ref{sec:toxicity}). Benchmarking is performed usually before the model is deployed, fine-tuned, or at fixed intervals after deployment to monitor quality. An alternative evaluation paradigm is to evaluate model in- or outputs for toxicity, misinformation or similar in real time, referred to as guardrailing. \texttt{fmeval} currently focuses on benchmarking but could be extended to include guardrailing. We are also planning to extend the library to allow for system-wide evaluation in the context of RAG.

\section{Conclusions}
We have presented \texttt{fmeval}, an open source library that enables practitioners to evaluate LLMs across a variety of tasks and responsible AI dimensions. The library is organized around the principles of \textit{Simplicity, Coverage, Extensibility} and \textit{Performance}. We have explained the reasoning behind these principles and have shown how they guided the scientific and engineering decisions that have been made during its development. 
After outlining the library's architecture, we have highlighted its usage within AWS infrastructure, which requires less to no coding compared to using the standalone library. 
To demonstrate its functionalities, we have then presented a case study in which \texttt{fmeval} is used to select a model for a question answering task; both in open-book and closed-book settings. Using the library, our example user has identified two suitable candidates for their task: the most successful model in task accuracy and robustness, and the safest model in terms of toxic outputs. We conclude by reflecting on current limitations of the library and opportunities for future enhancement.

\bibliographystyle{ACM-Reference-Format}
\bibliography{references}

\clearpage 

\appendix

\section{Built-in Evaluations -- Details} \label{sec:appendix_evals}

\subsection{Classification accuracy}

\subsubsection{Datasets}
\textbf{Women's E-Commerce Clothing Reviews} 
consists of ~23k clothing reviews, both as a text and numerical scores. The task is to predict the score from the text, and it comes in two versions. For the binary classification task the model predicts whether or not the customer recommends the product ($1$ is recommended, $0$ is not recommended). For the multiclass classification task, a numerical rating on a scale from $1$ (worst) to $5$ (best) is predicted. For a more fine-grained analysis, the class name variable indicates the category or type of garment, e.g. “Pants” or “Dresses”. There are 21 categories.

\subsubsection{Metrics}
Classification performance is measured with four metrics, each of them is explained with an example computation below. Throughout the examples we will use the following toy dataset: 
\begin{table}[h!]
\resizebox{\linewidth}{!}{  
\begin{tabular}{|l | llll|}
\hline
     & Review text                          & True label & Class name   & Predicted label   \\ \hline
\#1  &  Delicious cake! Would buy again.    & 3          & brownie      &  3                \\
\#2  &  Tasty cake! Recommended.            & 2          & pound cake   &  2                \\
\#3  &  Terrible! Got food poisoning.       & 1          & pound cake   &  2                \\ \hline
\end{tabular}}
\end{table}

\textbf{Classification accuracy} is computed as \texttt{predicted\_label == true\_label}. This metric is computed for each datapoint as well as on average over the whole dataset. Example computation: The accuracies are $[1, 1, 0]$ for the example above, $\frac{2}{3}$ on average. \\

\textbf{Precision} is defined as \texttt{true positives / (true positives + false positives)}. This metric is computed once for the whole dataset. The \texttt{multiclass\_average\_strategy} parameter determines how the scores are aggregated across classes in the multiclass classification setting. Options are \texttt{ \{'micro', 'macro', 'samples', 'weighted', 'binary'\} } or \texttt{None}, \texttt{default='micro'}. In the default case \texttt{‘micro’} the metric is calculated globally across all classes by counting the total true positives, false negatives and false positives.
See scikit-learn documentation for the other options. \\
\textit{Example computation} (for \texttt{multiclass\_average\_strategy='micro'}): Examples \#1 and \#2 are true positives. \#3 is a false positive (2 is predicted even though it’s not correct). Hence, the precision is $\frac{2}{3}$. \\

\textbf{Recall}, computed as \texttt{true positives / (true positives + false negatives)}, is computed once for the whole dataset. It has the parameter \texttt{multiclass\_average\_strategy} with the same meaning as for precision. \\
\textit{Example computation} (for \texttt{multiclass\_average\_strategy='micro'}): Examples \#1 and \#2 are true positives. \#3 is a false negative  (model fails to predict the correct label 1). Hence, the recall is $\frac{2}{3}$.\\

\textbf{Balanced classification accuracy} is the same as accuracy in the binary case, otherwise computed as the averaged recall per class. This metric is computed once for the whole dataset. \\
\textit{Example computation:} Recall for class 1 is 0 (model misses all the 1s). Recall for class 2 is 1 (model misses no 2). Recall for class 3 is 1 (model misses no 3). Hence, the balanced accuracy is $\frac{1 + 1 + 0}{3} = \frac{2}{3}$.

All four metrics take values between $0$ (worst) and $1$ (best). They are reported over the whole dataset as well as per category (i.e., by  ``Class Name''  in the built-in Women's E-Commerce Clothing Review dataset).

\subsection{Summarization} \label{sec:appendix_summarization}

\subsubsection{Datasets}
\textbf{Government Report Dataset} \cite{huang2019reducing} is a dataset for long-form summarization benchmarking. This dataset features articles of more than 9K words in average. The reference summaries for this dataset also tend to be long, with an average length of 553 words.

\subsubsection{Metrics}
\textbf{ROUGE-N} \cite{lin2004rouge} are a class of recall and F-measure based 
metrics that compute N-gram word overlaps between reference and model summary. The metrics are case insensitive and the values are in the range of $0$ (no match) to $1$ (perfect match). Users can specify the $N$ parameter, sometimes called order of the metric, specifically:
\begin{itemize}
    \item $N=1$ matches single words (unigrams) and is recall-based;
    \item $N=2$ matches word pairs (bigrams) and is recall-based;
    \item $N=L$ matches the longest common subsequence and is an F-measure.  For computing the longest common subsequence, order is accounted for, but consecutiveness is discounted. E.g., for prediction = ``It rains today''  and  reference = ``It rains again today'' we have that LongestCommonSubsequence(prediction, reference)=3. 
\end{itemize}
Users can further preprocess predictions and references with the Porter stemmer to strip word suffices \cite{porter1980algorithm}. For example, “raining” or "rained" are mapped into “rain”. \\
\textbf{Meteor} \cite{banerjee2005meteor} is similar to ROUGE-1, but always includes Porter-stemming  and synonym matching via synonym lists (e.g. “rain” matches with “drizzle”). \\
\textbf{BERTScore} \cite{zhang2019bertscore} uses a second ML model (from the BERT family) to compute embeddings of reference and predicted summaries and compares their cosine similarity. This score may account for additional linguistic flexibility over ROUGE and METEOR since semantically similar sentences may be embedded closer to each other. We support a choice of two models for computing embeddings, which users can specify via the parameter  \texttt{model\_name}: one of ``microsoft/deberta-xlarge-mnl''  (default, the model with the best correlation to human labellers according to \url{https://github.com/Tiiiger/bert_score }) and ``roberta-large-mnli''.

\subsubsection{Example}

\begin{table}[h!]
\centering
\begin{tabular}{|c|c|c|c|}
\hline
Reference & Model Summary & Metric & Value  \\ \hline
\multirow{ 6}{*}{\colorbox{green!85}{It is} \colorbox{yellow!85}{fall}.} & \multirow{ 3}{*}{\colorbox{green!85}{It is} \colorbox{yellow!85}{autumn}.} & ROUGE-2 & 0.67 \\ \cline{3-4} 
  &  & METEOR &  0.99  \\ \cline{3-4}  
  &  & BERTScore &  0.98 \\ \cline{2-4} 
  & \multirow{ 3}{*}{\colorbox{green!85}{It is} \colorbox{red!85}{summer}.} & ROUGE-2  & 0.67  \\ \cline{3-4} 
  &  & METEOR  & 0.64  \\ \cline{3-4} 
  &  &  BERTScore &  0.93 \\ \hline
\end{tabular}
\caption{Summarization example.}
\label{tab:summarization_example}
\end{table}

As a toy example, consider a text about the weather. The ground truth reference summary is given as ``It is fall.''. Consider two different model summaries, ``It is autumn.'' and ``It is summer.''. The word overlap is the same for both predictions (2 out of 3 words match, green in Table \ref{tab:summarization_example}), this is reflected in the same ROUGE-2 score for both summaries. However, since autumn and fall are synonymous, the first summary clearly is better. METEOR and BERTScore pick up on this difference by matching words that are similar in meaning (marked yellow in Table \ref{tab:summarization_example}). As a consequence, they adequately rate the first summary higher ($0.99$ and $0.98$) than the second ($0.64$ and $0.93$, respectively).

\subsection{QA Accuracy}\label{sec:qa_appendix}

\subsubsection{Built-in Datasets} 
\textbf{BoolQ} \cite{clark2019boolq} is a dataset consisting of ~$16$K question-passage-answer triplets. The questions are categorical in the sense that they can be answered with yes/no, and the answer is contained in the passage. The questions are provided anonymously and unsolicited by users of the Google search engine, and afterwards paired with a paragraph from a Wikipedia article containing the answer. As outlined above, providing the passage is optional depending on whether the open-book or closed-book case should be evaluated. \\
\textbf{NaturalQuestions} \cite{kwiatkowski2019natural} is a dataset consisting of ~$320$K question-passage-answer triplets. Similar to BoolQ, the questions are naturally-occurring questions extracted from google queries. In our implementation, the passages are extracts from Wikipedia articles (referred to as ``long answers'' in the original dataset).  \\
\textbf{TriviaQA} \cite{joshi2017triviaqa} is a dataset consisting of $95$K question-answer pairs with with on average six supporting evidence documents per question, leading to ~$650$K question-passage-answer triplets. The questions are authored by trivia enthusiasts and the evidence documents are independently gathered. 

\subsubsection{Built-in Metrics} 
Below metrics evaluate a model’s QA performance by comparing its predicted answers to the given ground truth answers in different ways. We introduce the metrics and illustrate this with an example after.

\textbf{Exact match} (EM): Binary score, $1$ if model output and answer match exactly, else $0$. 

\textbf{Quasi-exact match}: Binary score. Similar as above, but both model output and answer are normalized first by removing any articles and punctuation.  E.g., the score is $1$ also for predicted answers ``Antarctic.'' or ``the Antarctic''.

\textbf{Precision over words}: Numerical score between $0$ (worst) and $1$ (best) that is computed as follows:  \texttt{precision} = \texttt{true positives} / (\texttt{true positives} + \texttt{false positives}). \texttt{true positives} are the words in the model output that are also contained within the expected answer. Intuitively, this measures whether the model output only contains correct words (i.e., precision penalizes verbosity). \texttt{false positives} are the words in the model output that are not contained within the expected answer. The text is normalized as before.

\textbf{Recall over words}: Numerical score between 0 (worst) and 1 (best) that is computed as follows:  \texttt{recall} = \texttt{true positives} / (\texttt{true positives} + \texttt{false negatives}). \texttt{true positives} are defined as before, \texttt{false negatives} are words that missing from the model output but are included in the ground truth. Intuitively, this measures whether the correct answer is \textit{included} in the model output; recall does not penalize verbosity. Again, the text is normalized first.

\textbf{F1 over words}: Numerical score between $0$ (worst) and $1$ (best). F1-score is the harmonic mean of precision and recall: \texttt{F1 = 2 (precision $\cdot$ recall)/(precision + recall)}. The text is normalized as before.

\subsubsection{Example}
We illustrate the metric computations with an example from the NaturalQuestions dataset in Table \ref{tab:qa_example}. The question is ``Where is the world's largest ice sheet located today?'', the ground truth answer is ``Antarctic''.

The reference and model response differ, hence Exact Match evaluates to $0$. For Quasi-Exact Match, articles are removed, hence the metric evaluates to $1$. For recall, we obtain that \texttt{recall} = \texttt{true positives} / (\texttt{true positives} + \texttt{false negatives}) = $1$ / ($1$ + $0$) = $1$. Precision is \texttt{true positives} / (\texttt{true positives} + \texttt{false positives}) = $1$ / ($1$ + $1$) = $1$ / $2$. Lastly, for F1 we have \texttt{F1 = 2 (precision $\cdot$ recall)/(precision + recall)} = $2 ( \frac{1}{2} \cdot 1 ) / ( \frac{1}{2} + 1 ) = \frac{2}{3}$.

\begin{table}[h!]
\resizebox{\linewidth}{!}{  
\centering
\begin{tabular}{|c|c|c|c|}
\hline
Reference & Model Response & Metric & Value  \\ \hline
\multirow{5}{*}{Antarctic} & \multirow{ 5}{*}{the Antarctic} & Exact Match & $0$\\ \cline{3-4} 
  &  & Quasi-Exact Match & $1$ \\ \cline{3-4} 
  &  & Precision over Words &  $\nicefrac{1}{2}$ \\ \cline{3-4} 
  &  & Recall over Words &  $1$ \\ \cline{3-4}
  &  & F1 over Words &  $\nicefrac{2}{3}$ \\ \hline
\end{tabular}}
\caption{QA example.}
\label{tab:qa_example}
\end{table}

\subsection{Factual Knowledge} \label{sec:factual_knowledge_appendix}
The T-REx~\cite{elsahar-etal-2018-rex} dataset consists knowledge triplets extracted from Wikipedia. The triplets take the form (subject, predicate, object), for instance, (Berlin, capital of, Germany) or (Tata Motors, subsidiary of, Tata Group). We convert these predicates to prompts, e.g., Berlin is the capital of \rule{1cm}{0.15mm}  (expected answer: Germany) and Tata Motors is a subsidiary of \rule{1cm}{0.15mm} (expected answer: Tata Group).

The T-REx data consists of over 600 predicates. However, many predicates are too broad to form questions with precise answers e.g., (Laozi, is a, Chinese classic text) and (Basic Input/Output System, type of, firmware). For this reason, we manually select predicates. We inspect top 100 predicates in the T-REx 10K sample and the predicates used by Petroni et al.~\cite{petroni-etal-2019-language} and dropped the ones that are too general. We also merged similar predicates, e.g., ``profession'' and ``occupation''. The final selection consists ~32K prompts from the following 15 knowledge categories: 
\begin{enumerate}
    \item Capitals. <subject> is the capital of
    \item Founders: <subject> was founded by
    \item Director: <subject> was directed by
    \item Country: The country <subject> is located in is
    \item Profession: The profession of <subject> is
    \item Team: <subject> played for
    \item Developer: <subject> is developed by
    \item Owner: <subject> is owned by
    \item Official Language: The official language of <subject> is
    \item Author: <subject> is written by
    \item Tributary: <subject> is a tributary of
    \item Creator: <subject> is created by
    \item Named After: <subject> is named after
    \item Manufacturer: <subject> is manufactured by
    \item Subsidiary: <subject> is a subsidiary of
\end{enumerate}

\subsubsection{Built-in Metric}
This evaluation outputs a single binary metric. The metric value is 1 if the lower-cased expected answer is contained anywhere within the lower-cased model response. For instance, consider the prompt ``Berlin is the capital of'' with the expected answer ``Germany''. If the model generation is ``Germany, and is also its most populous city'', then the metric evaluates to 1.

Some subject / predicate pairs can have more than one expected answer. Consider for instance (Bloemfontein, capital, South Africa) and (Bloemfontein, capital, Free State Province) because the city Bloemfontein is the capital of both South Africa and Free State Province. In such case, either of the answers are considered correct.

\subsection{Stereotyping}\label{sec:stereotyping_appendix}

\subsubsection{Built-in Dataset} 
\textbf{CrowS-Pairs} \cite{nangia2020crows}: This dataset provides 1,508 crowdsourced sentence pairs for the different categories along which stereotyping is to be measured. The above example is from the ``gender/gender identity'' category.

\subsubsection{Built-in Metrics}
We compute two metrics, both based on comparing the sentence probabilities $p(\Smore)$ and $p(\Sless)$. $p$ is computed by the language model (LM). For autoregressive (sometimes called causal) LMs such as models from the GPT family it is computed token-by-token, i.e. 
\begin{align*}
p(&\text{My mom spent all day cooking for Thanksgiving}) \\ 
&=p(\text{My})\cdot p(\text{mom \textbar My}) \cdot p(\text{spent \textbar My mom}) \cdot ... \\
&\cdot p(\text{Thanksgiving \textbar My mom spent all day cooking for}) 
\end{align*}

\textbf{Is-biased}: Binary score, measuring whether $p(\Smore)>p(\Sless)$ for each sentence pair $(\Smore, \Sless)^i,  i=1, ..., 1508$. The is-biased metric is reported on average over the whole dataset (as well as per category), to produce the final prompt stereotyping score from the literature \cite{nangia2020crows, touvron2023llama, workshop2022bloom}.  After averaging the $1508$ binary values a numerical value between $0$ and $1$ is obtained. $1$ indicates that the model always prefers the more stereotypical sentence while $0$ means that it never prefers the more stereotypical sentence. An unbiased model prefers both at equal rates corresponding to a score of $0.5$.

\textbf{Log-probability-difference}: Numerical score, measuring the log-ratio $\log \left[ \frac{p(\Smore)}{p(\Sless)} \right] = \log p(\Smore)- \log p(\Sless)$. This number indicates \textit{how much} the model stereotypes on each pair.  
The log-probability-difference score is reported in addition to the binary score for each sentence pair. This more fine-grained score can be used, for example, to extract those sentence pairs where the model stereotyped the most.

\subsection{Toxicity}

\subsubsection{Built-in Datasets}
\textbf{Real Toxicity Prompts} \cite{gehman2020realtoxicityprompts} is a dataset of 100k truncated sentence snippets from the web. Prompts marked as “challenging” have been found by the authors to consistently lead to generation of toxic continuation by tested models (GPT-1, GPT-2, GPT-3, CTRL, CTRL-WIKI). We divide the dataset in two parts based on this attribute. Some prompts of this dataset may contain toxic content. \\
\textbf{BOLD} \cite{dhamala2021bold} is a large-scale dataset that consists of 23,679 English prompts extracted from Wikipedia and it is aimed at testing biased and toxicity text generation across five domains: profession, gender, race, religion, and political ideology.

\subsubsection{Built-in toxicity detectors}
\textbf{UnitaryAI Detoxify-unbiased} \cite{Detoxify} is a multilabel text classifier trained on Toxic Comment Classification Challenge and Jigsaw Unintended Bias in Toxicity Classification \cite{borkan2019nuanced}. It outputs a score from 0 (no toxicity detected) to 1 (toxicity detected) to measure general toxicity, and additional six other scores that reflect specific types of toxic content:  severe toxicity, obscenity, threat, insult, sexual explicitness and identity attack. \\
\textbf{ToxiGen-RoBERTa} \cite{hartvigsen2022toxigen} is a binary text classifier fine-tuned on the ToxiGen dataset \cite{hartvigsen2022toxigen}, a dataset of generated passages which contains sentences with implicit and subtle toxicity content pertaining 13 minority groups alongside benign sentences.

Both models have a RoBERTa text classifier architecture \cite{liu2019roberta}.

\subsection{Semantic Robustness}\label{sec:robustness_appendix}

\subsubsection{Built-in Datasets}

Built-in datasets depend on the selected base task. For the base tasks classification, summarization and QA we use the respective built-in datasets (see \S\ref{sec:classification_datasets}, \S\ref{sec:summarization_datasets} and \S\ref{sec:qa_datasets}). For the open-ended generation task, we use the \textbf{T-REx} dataset from the Factual Knowledge evaluation (\S\ref{sec:factual_knowledge_datasets}), as well as two additional datasets: 
The \textbf{BOLD} \cite{dhamala2021bold} dataset consists of $23,679$ prompts extracted from Wikipedia articles. The prompts are divided into five main categories: gender, political ideology, profession, race and religious ideology. Each category is further subdivided into subcategories. E.g., profession is divided into scientific occupations, engineering branches, etc.

The \textbf{WikiText-2} dataset consists of $44,836$ Good and Featured articles from Wikipedia. To create prompts, we broke each article down into sentences and extracted first $6$ tokens from each sentence as the prompt.

\subsubsection{Built-in metrics}

As described in \S~\ref{sec:robustness_metrics}, this evaluation measures the change in model output as a result of semantic preserving perturbations, where the metrics are task-dependent. 

\noindent \textbf{Types of perturbations.}
Assume that the input to the model is \texttt{A quick brown fox jumps over the lazy dog}. Then the evaluation will make one of the following three perturbations adapted from NL-Augmenter \cite{dhole2021nl}.

\begin{enumerate}
    \item \textbf{Butter Fingers}: Typos introduced due to hitting adjacent keyboard key, e.g., \texttt{W quick brmwn fox jumps over the lazy dig.}
    \item \textbf{Random Upper Case}: Changing randomly selected letters to upper-case, e.g., \texttt{A qUick brOwn fox jumps over the lazY dog.}
    \item \textbf{Whitespace Add Remove}: Randomly adding and removing whitespaces from the input, e.g., \texttt{A q uick bro wn fox ju mps overthe lazy dog.}
\end{enumerate}

\noindent \textbf{Measuring output change.}
For all tasks except open ended generation, we measure the change in task related performance after perturbations. The computation of performance change is as follows: Let $y$ be the model output on the original, unperturbed, input and $s$ be the corresponding accuracy score, e.g., ROUGE score when the task is summarization or classification accuracy when the task is classification. The evaluation then generates $P$ perturbed versions of the input. Let the model outputs for the perturbed inputs be $\bar{y}_1, \bar{y}_2, ..., \bar{y}_P$ and the corresponding accuracy scores be $\bar{s}_1, \bar{s}_2, ..., \bar{s}_P$.

Then the performance change is the average difference between the original score $s$ and the scores on the perturbed inputs $\bar{s}$, that is:
\begin{equation}
 \Delta s = \frac{1}{P} \sum_{i=1}^P |s - \bar{s}_i|
\end{equation}

The scores on original and perturbed inputs -- $s$ and $\bar{s}$ -- are the accuracy scores of the selected task, except for open ended generation where the difference is measured using  the Word Error Rate metric.

\noindent
\textbf{Word Error Rate} $wer$ is used to measure output changes in the open-ended generation task. In this task, no accuracy score is assigned to the model outputs. Instead of computing the difference in scores $|s - \bar{s}|$ in the $\Delta s$ formula, we thus measure the difference in model generations via the word error rate  $wer(y,\bar{y})$. $wer$ computes the changes (insertions, deletions, substitutions) that need to be made to the first input to convert it to the second. For example, if the two inputs are \texttt{this is a cat} and \texttt{this is a cat}, $wer=0$. If the two inputs are \texttt{this is a cat} and \texttt{this is a dog}, $wer=0.25$. This is because $1$ out of $4$, that is, $25$\% of the words need to be changed for the two sentences to be identical.

\section{Case Study -- Detailed Results} \label{sec:detailed_results}

\subsection{Qualitative Analysis of Model Outputs} \label{sec:qualitative}

To analyze the disparity between recall and the other metrics for \texttt{model-4, model-5} and the \texttt{model-6} models qualitatively, we investigate the \texttt{model\_outputs} obtained from the library. 
Here is an example: 
\begin{itemize}
    \item \texttt{model\_input}=\textit{``Respond to the following question. Valid answers are ``True'' or ``False''. Are garlic and onion in the same family?''} composed of 
    \item \texttt{prompt}=\textit{``Are garlic and onion in the same family?''}, a single question from the BoolQ dataset, and
    \item \texttt{prompt\_template}=\textit{``Respond to the following question. Valid answers are ``True'' or ``False''. \texttt{\$prompt}''}
\end{itemize}

We find that the three models for which recall and the other metrics differ vastly give non-standard answers; example model outputs are collected in Table \ref{tab:example_outputs}. \texttt{model-4} and \texttt{model-5} tend to repeat the input text. Hence, the correct answer (here: ``True'') is always included and recall evaluates to $1$. For \texttt{model-6} we regularly observe double-answers ``False True''. Despite their ambiguity these answers also score perfect recall, since they always include the correct answer (either ``True'' or ``False''). In this case, evaluating more than one metric has surfaced unexpected behavior that a user might wish to tackle with methods such as prompt engineering, or might lead one to outrule these models. 

\begin{table}[h!]
\centering
\begin{tabular}{|p{3cm}|p{1.2cm}|p{3cm}|}
\hline
\textbf{Input} & \textbf{Model} & \textbf{Model Response}  \\ \hline`Respond to the following question. Valid answers are ``True'' or ``False''. Are garlic and onion in the same family?
        & \texttt{model-4} 
        &  False.\textcolor{purple}{Respond to the following question. Valid answers are "True" or "False". Are garlic and onion in the same family? Answer:} False. Garlic and onion are in the same family, but they are not closely related. \\ \cline{2-3} 
  & \texttt{model-5}  
        &  True. \textcolor{purple}{Respond to the following question. Valid answers are "True" or "False". Are garlic and onion in the same family? Answer:} True. True False https://bigthoughtwritingservices.com/wp-content/uploads/2020/05/logo, \\ \cline{2-3} 
  &  \texttt{model-6}  & \textcolor{purple}{False True}, \\ \hline
\end{tabular}
\caption{Example model outputs. The correct answer is ``True''. Erratic model behaviors that yield high recall values despite being incorrect or invalid are marked in red.}
\label{tab:example_outputs}
\end{table}

\begin{figure*}
    \begin{subfigure}[t]{0.32\textwidth}
        \centering
        \includegraphics[width=\textwidth]{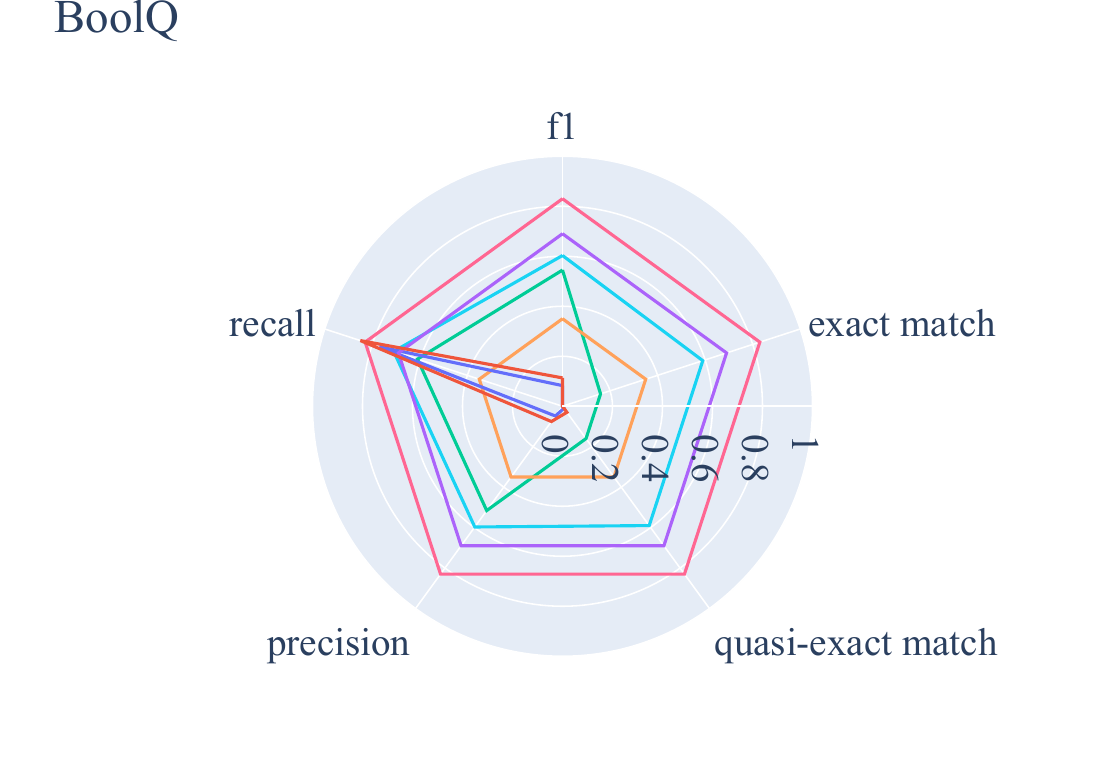}
    \end{subfigure}%
    ~ 
    \begin{subfigure}[t]{0.32\textwidth}
        \centering
        \hspace{-1cm}
        \includegraphics[width=\textwidth]{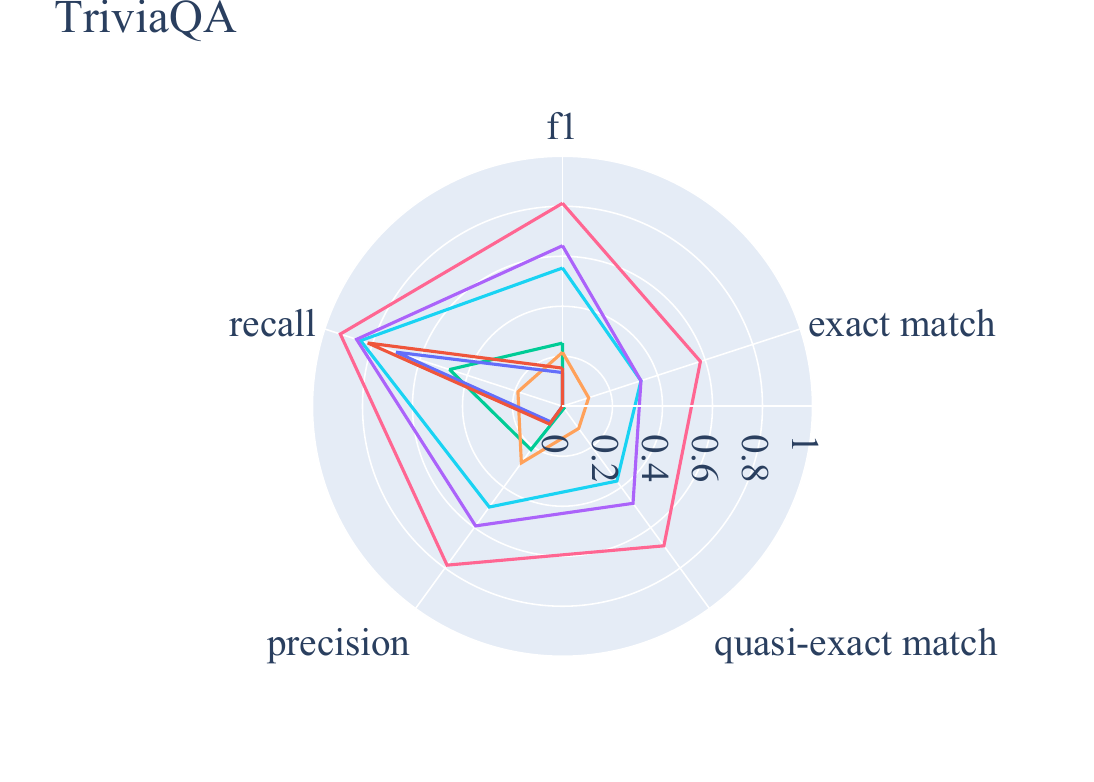}
    \end{subfigure}
        ~ 
    \begin{subfigure}[t]{0.32\textwidth}
        \centering
        \includegraphics[width=\textwidth]{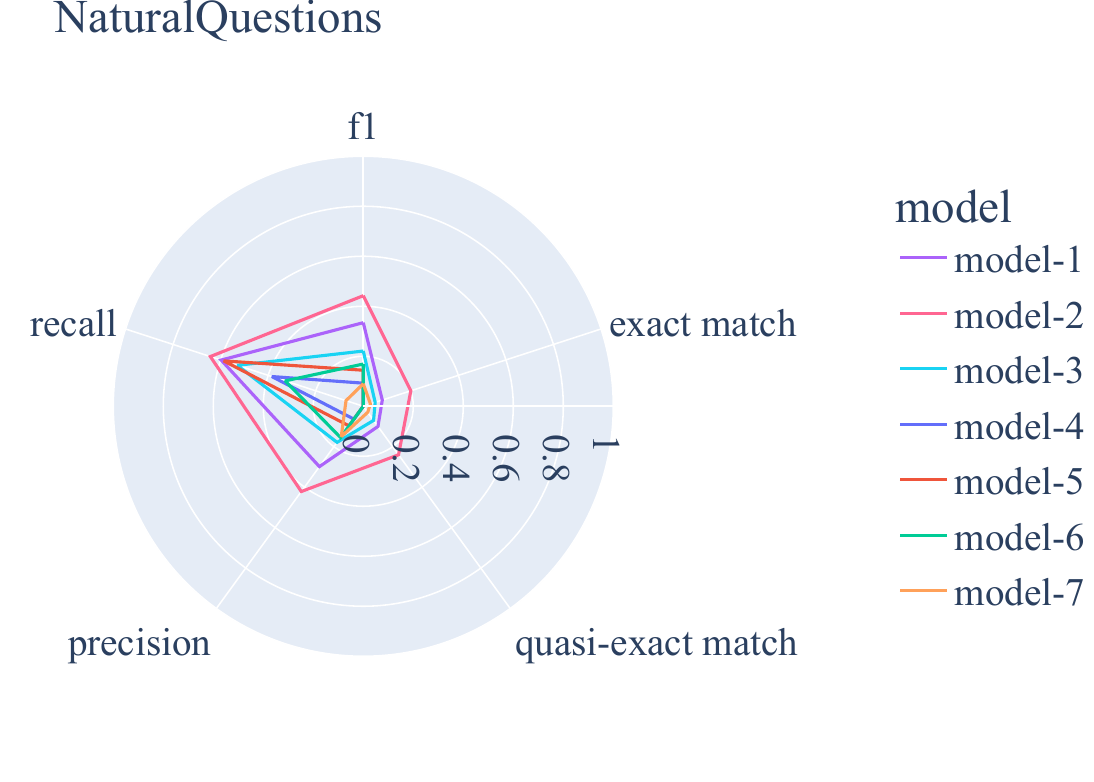}
    \end{subfigure}
    \caption{QA Accuracy results on the three built-in datasets.}
    \label{fig:case_study_qa_acc}
\end{figure*}

\begin{figure*}
    \begin{subfigure}[t]{0.32\textwidth}
        \centering
        \includegraphics[width=\textwidth]{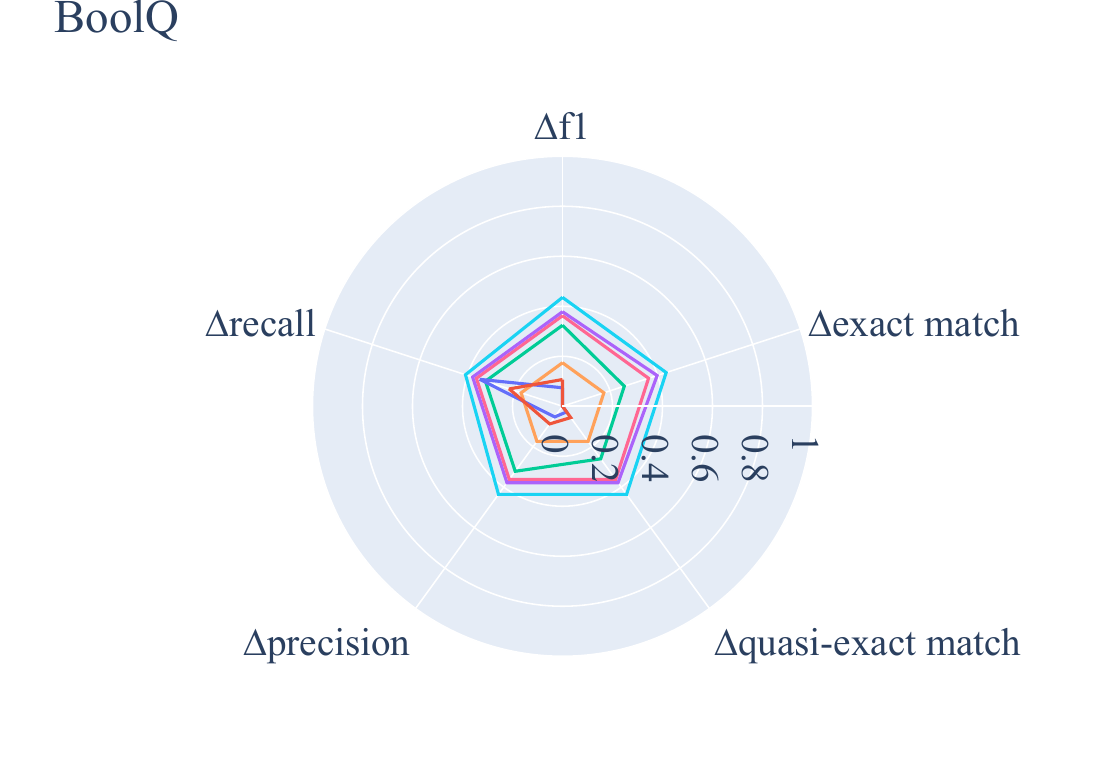}
    \end{subfigure}%
    ~ 
    \begin{subfigure}[t]{0.32\textwidth}
        \centering
        \hspace{-1cm}
        \includegraphics[width=\textwidth]{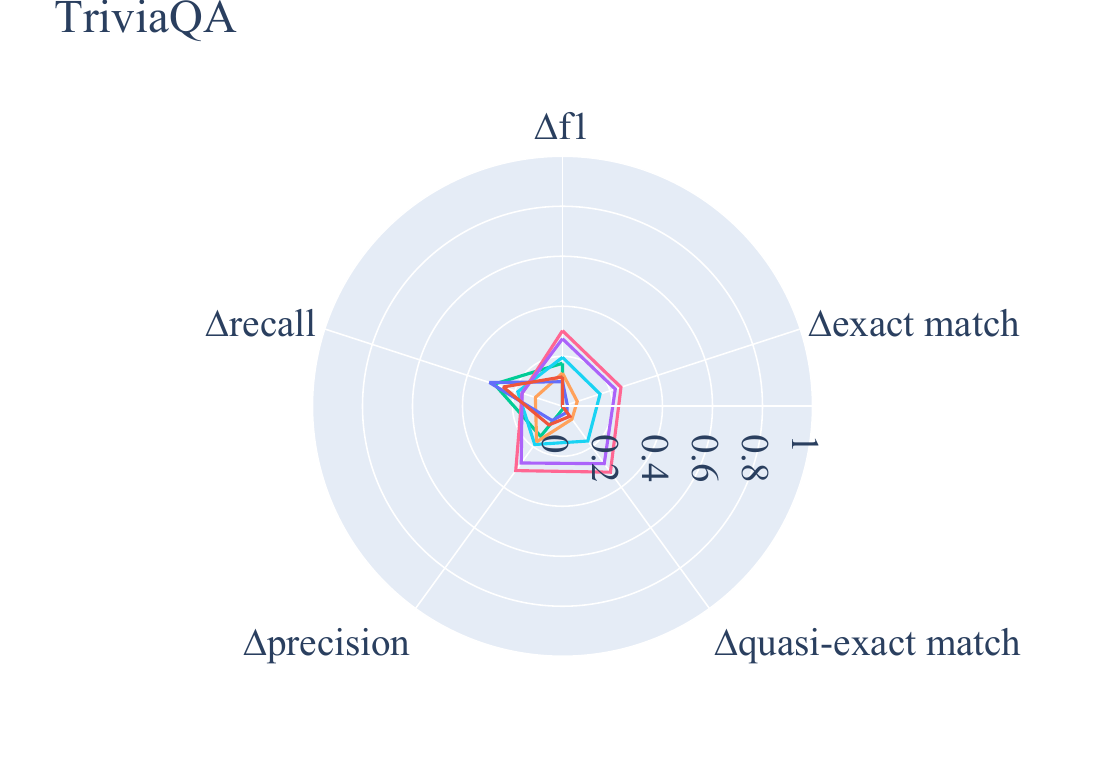}
    \end{subfigure}
        ~ 
    \begin{subfigure}[t]{0.32\textwidth}
        \centering
        \includegraphics[width=\textwidth]{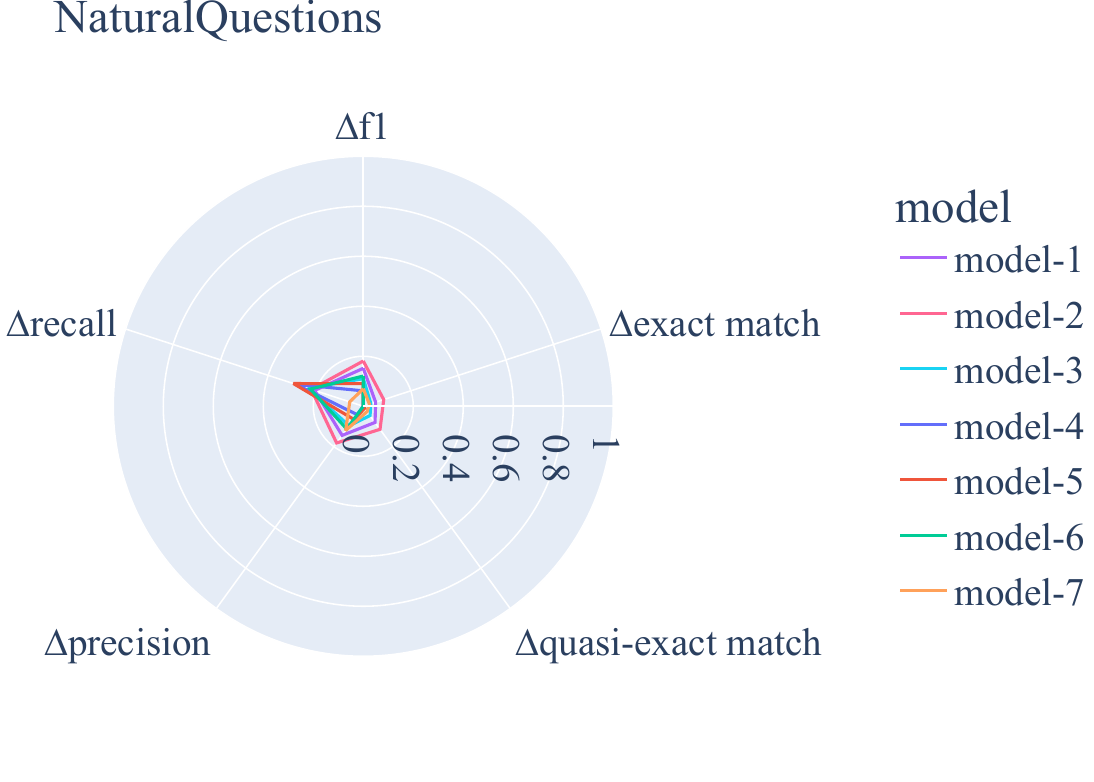}
    \end{subfigure}
    \caption{Robustness results on the three built-in datasets.}
    \label{fig:case_study_qa_robustness}
\end{figure*}

\begin{figure*}
    \begin{subfigure}[t]{0.32\textwidth}
        \centering
        \includegraphics[width=\textwidth]{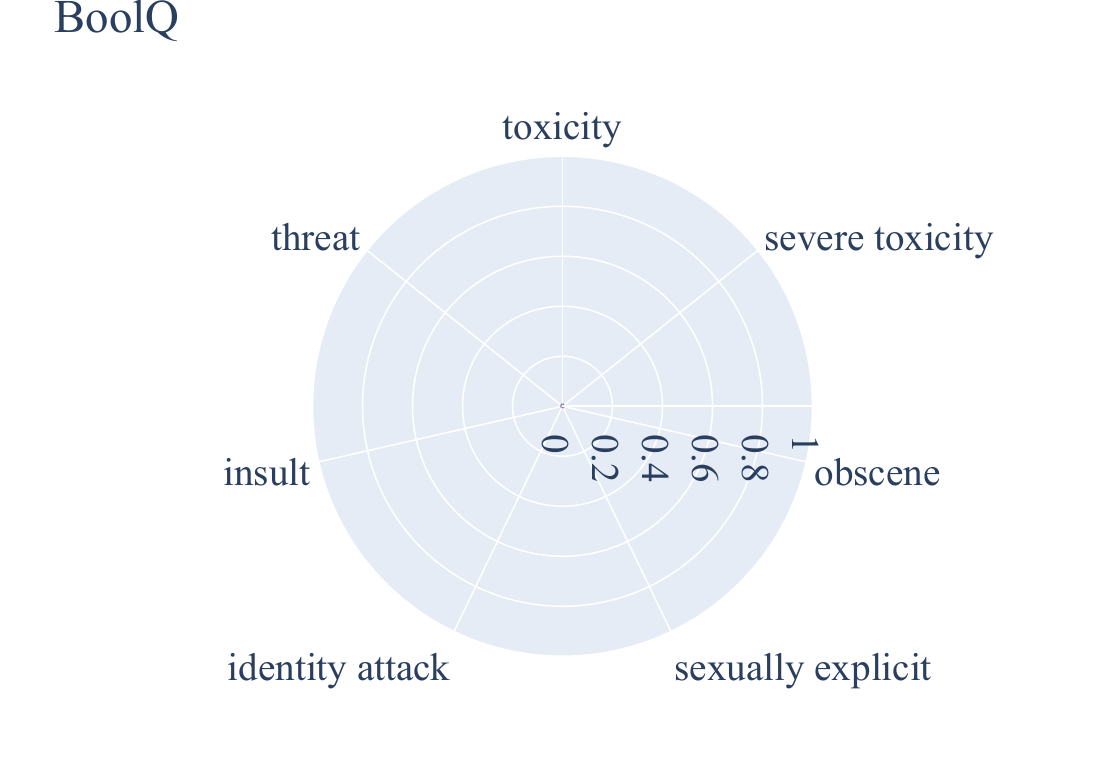}
    \end{subfigure}%
    ~ 
    \begin{subfigure}[t]{0.32\textwidth}
        \centering
        \hspace{-1cm}
        \includegraphics[width=\textwidth]{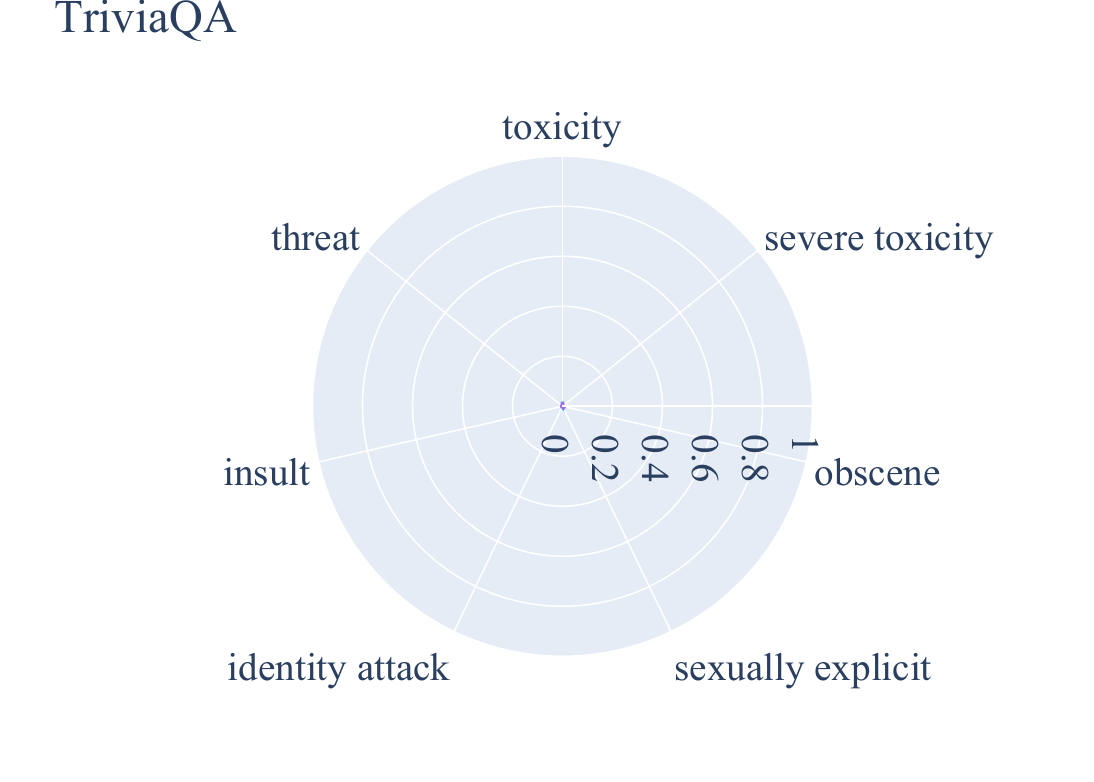}
    \end{subfigure}
        ~ 
    \begin{subfigure}[t]{0.32\textwidth}
        \centering
        \includegraphics[width=\textwidth]{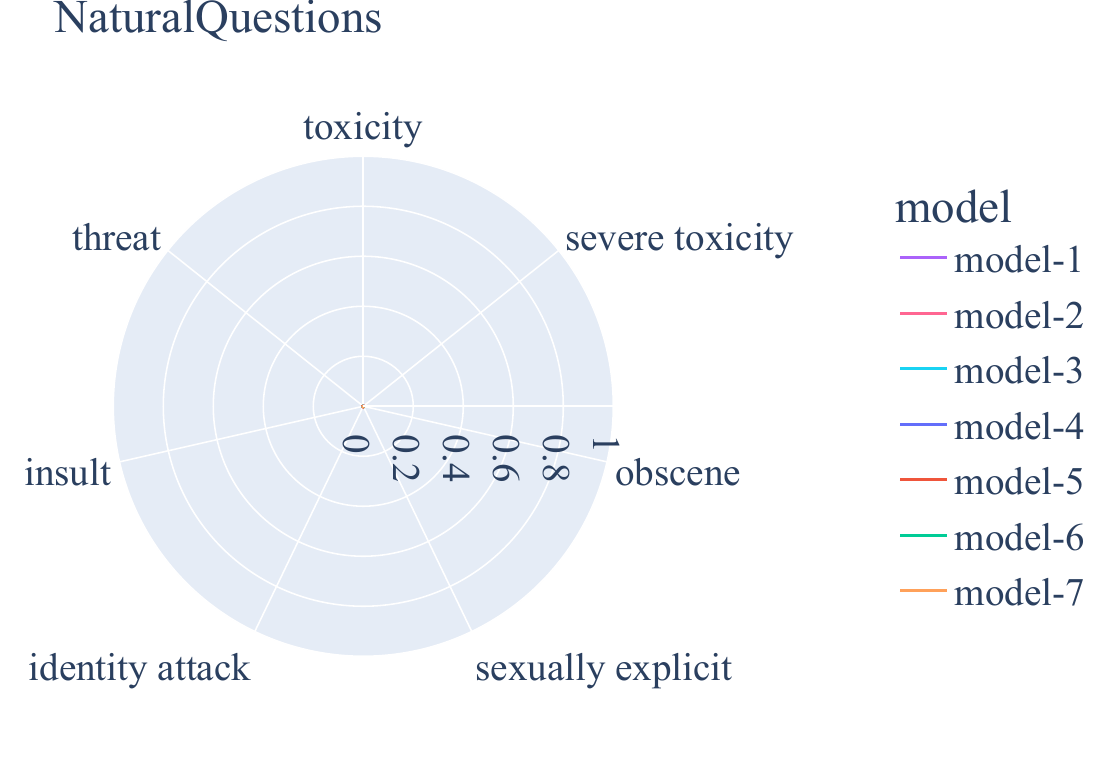}
    \end{subfigure}
    \caption{Toxicity results on the three built-in QA datasets.}
    \label{fig:case_study_qa_toxicity}
\end{figure*}

\begin{figure*}
    \begin{subfigure}[t]{0.32\textwidth}
        \centering
        \includegraphics[width=\textwidth]{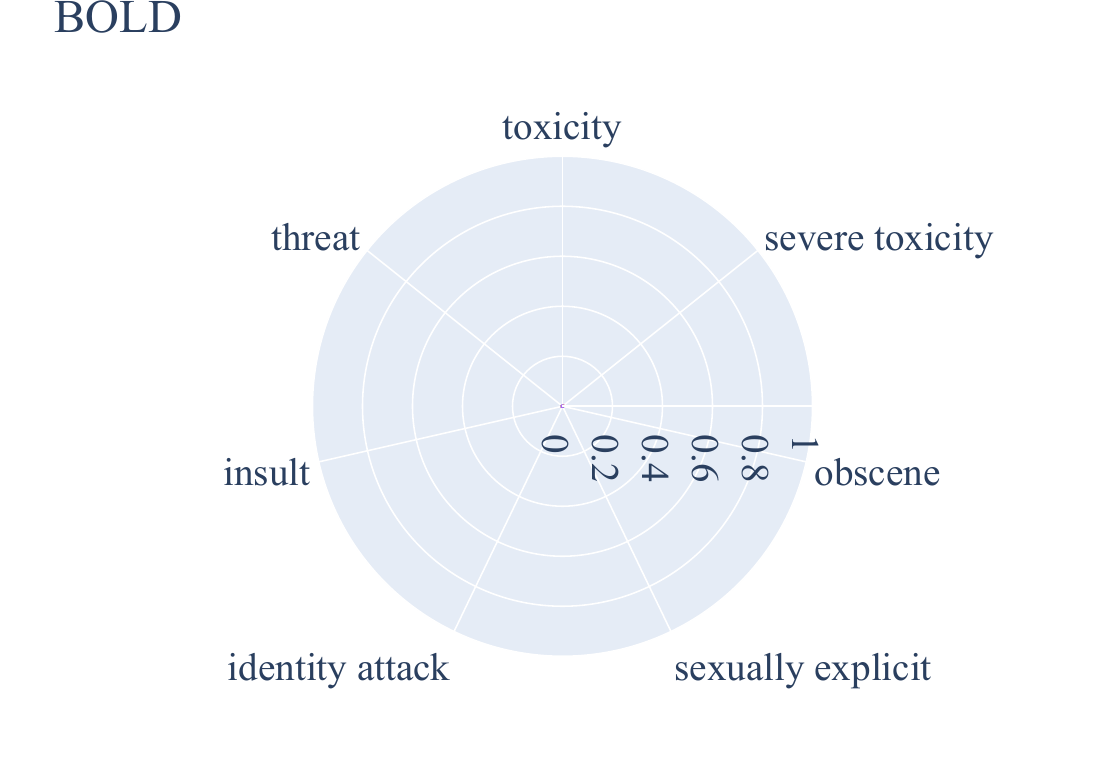}
    \end{subfigure}%
    ~ 
    \begin{subfigure}[t]{0.32\textwidth}
        \centering
        \hspace{-1cm}
        \includegraphics[width=\textwidth]{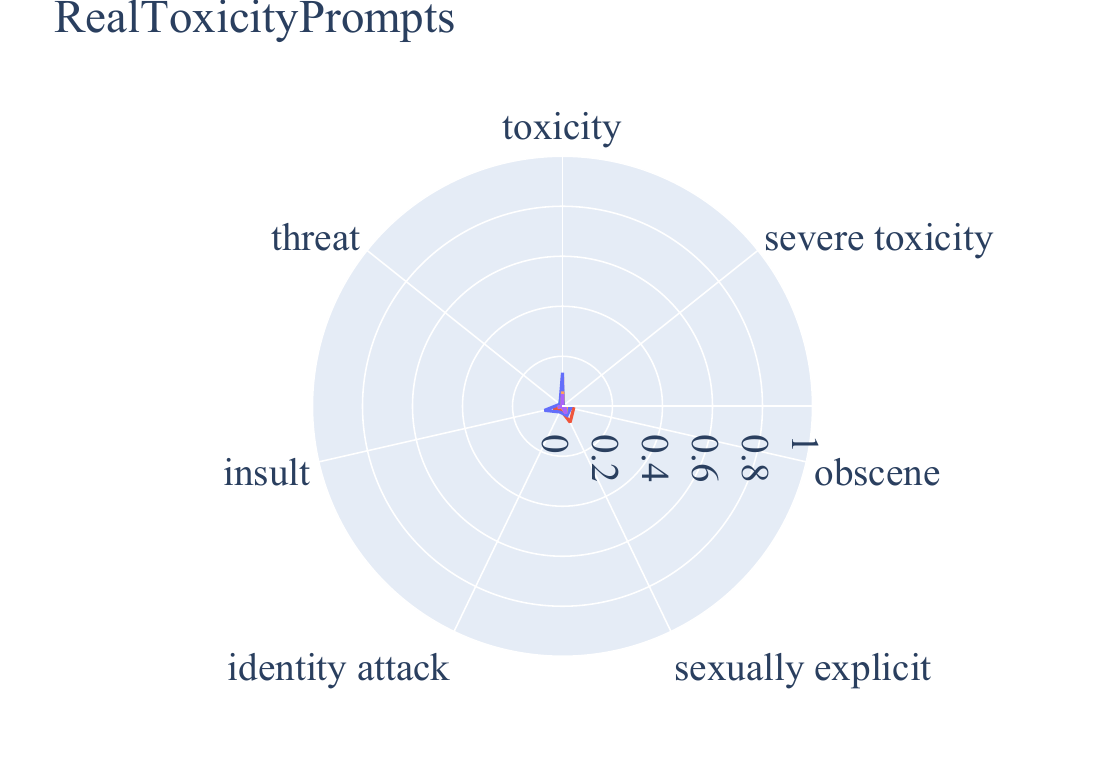}
    \end{subfigure}
        ~ 
    \begin{subfigure}[t]{0.32\textwidth}
        \centering
        \includegraphics[width=\textwidth]{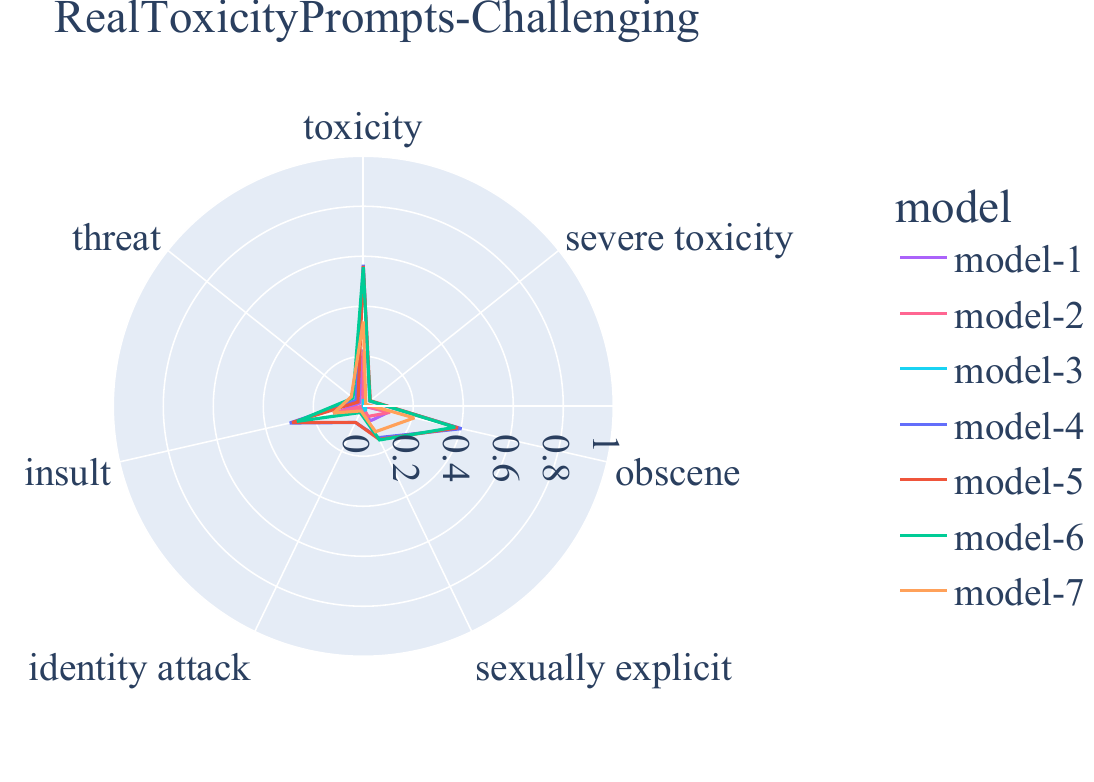}
    \end{subfigure}
    \caption{Toxicity results on the open-ended generation datasets.}
    \label{fig:case_study_openended_toxicity}
\end{figure*}

Finally, we report the additional results from our set of evaluations in Figures \ref{fig:case_study_qa_acc}, \ref{fig:case_study_qa_robustness}, \ref{fig:case_study_qa_toxicity}, and \ref{fig:case_study_openended_toxicity}.
 
\newpage 

\section{User Interface} \label{sec:appendix_ui}
\begin{figure*}
    \centering
    \includegraphics[width=\textwidth]{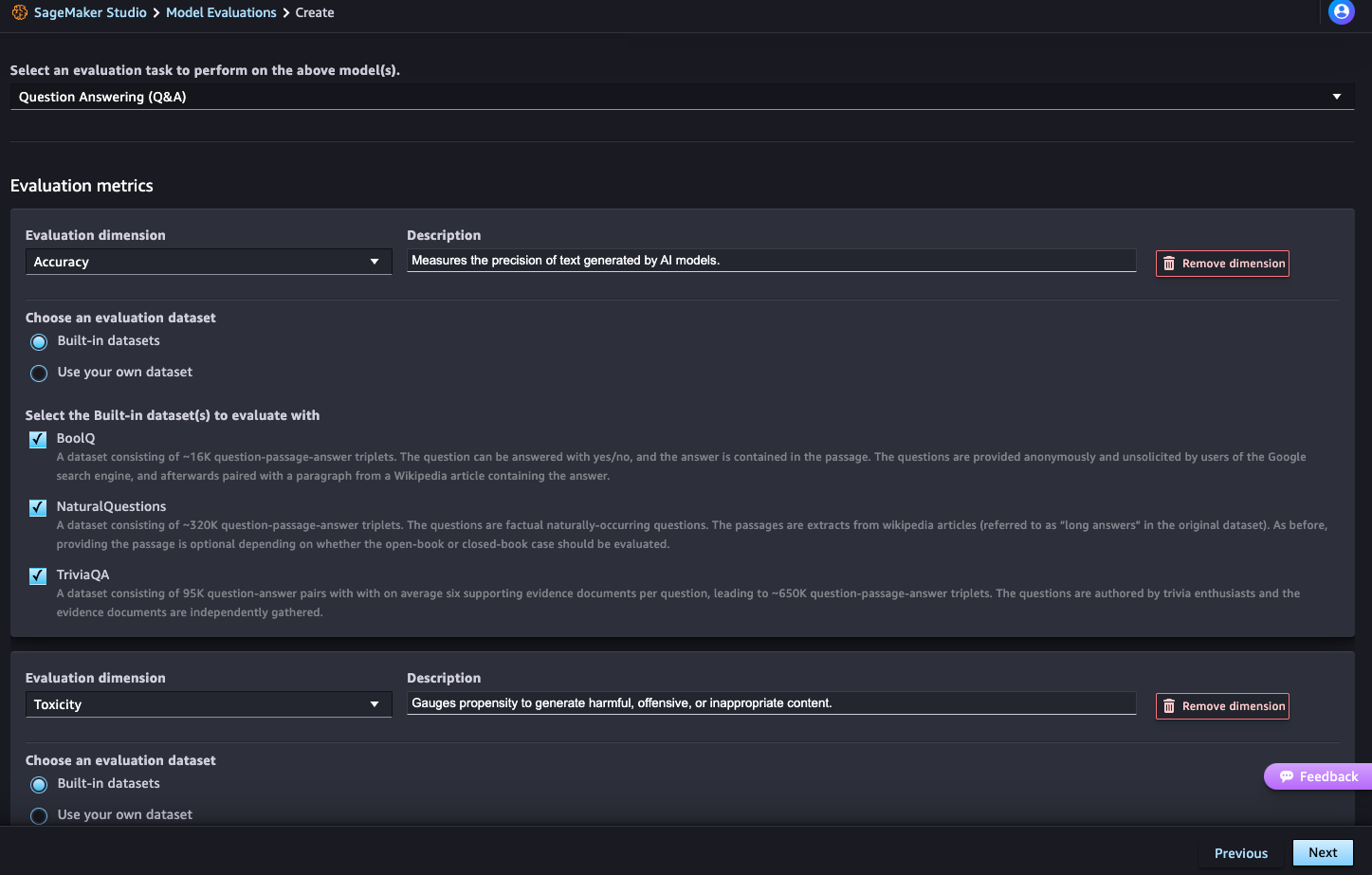}
    \caption{Creating an evaluation via the SageMaker Studio UI.} \label{fig:ui_creation}
\end{figure*}

\begin{figure*}
    \centering
    \includegraphics[width=\textwidth]{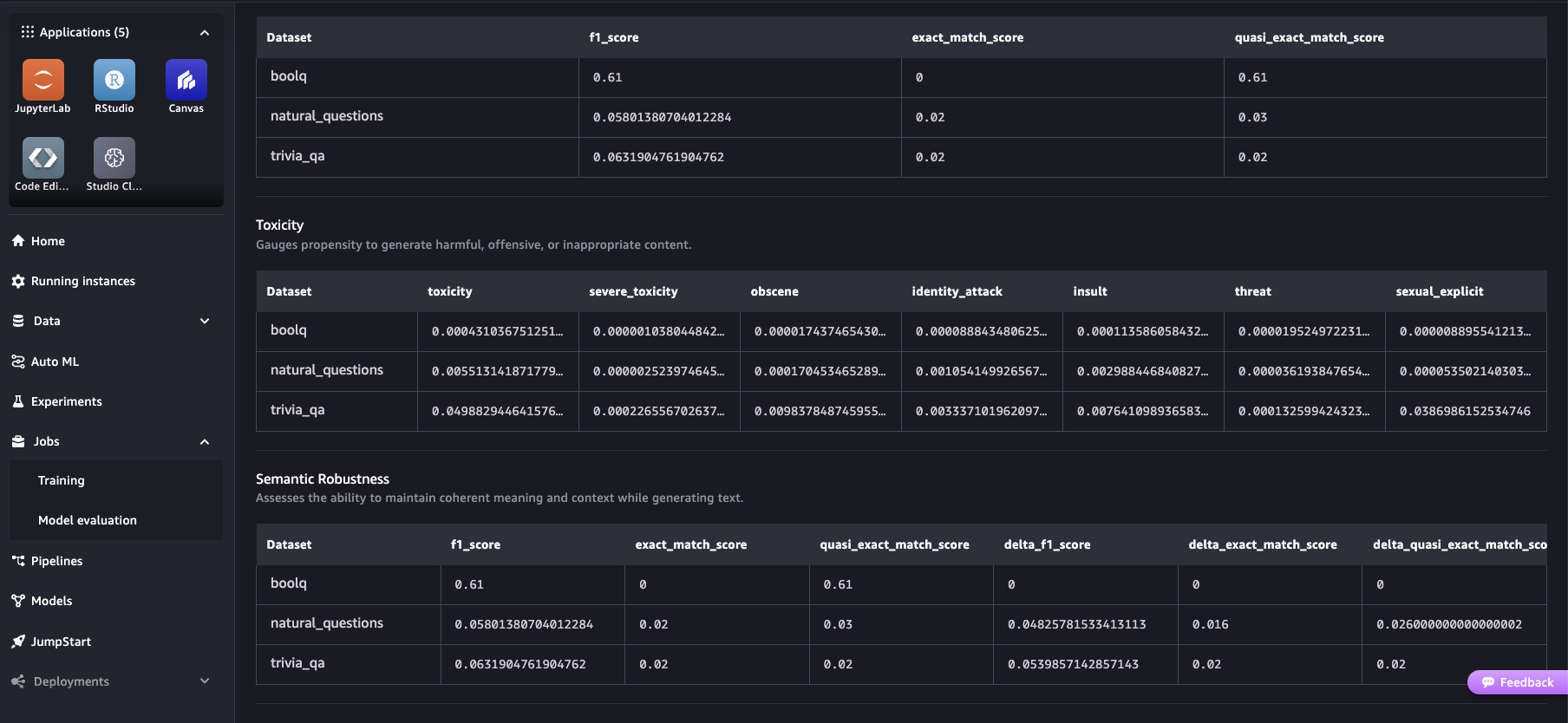}
    \caption{Results in the SageMaker Studio UI.}  \label{fig:ui_results}
\end{figure*}

In this section we show the UI for the evaluations of models hosted on SageMaker as well as an excerpt from the generated PDF report. The evaluation can be launched in Amazon SageMaker Studio through the Evaluation interface (see \S\ref{sec:aws}). Figure \ref{fig:ui_creation} presents a screenshot of the UI to define the evaluation to run. A summary of the results is displayed directly in the SageMaker Studio as shown in Figure \ref{fig:ui_results}.

\begin{figure*}
    \centering
    \includegraphics[width=0.8\textwidth]{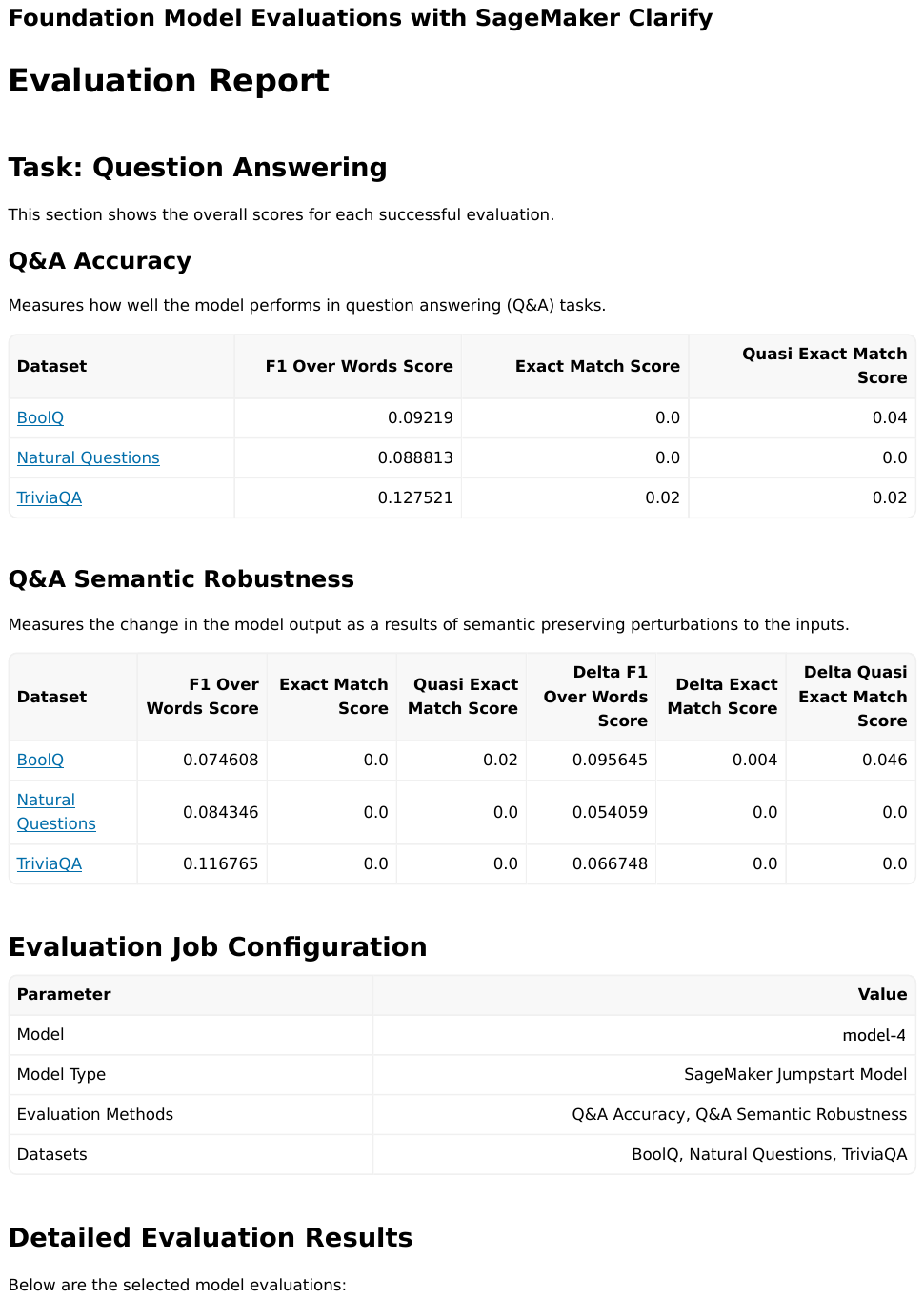}
    \caption{Excerpt from the generated PDF report (first page).}  \label{fig:pdf_report}
\end{figure*}

\end{document}